\pdfoutput=1
\documentclass[11pt]{article}
\usepackage{nodalida2025}
\usepackage{times}
\usepackage{url}
\usepackage{latexsym}

\aclfinalcopy
\usepackage[hidelinks]{hyperref}
\usepackage{pifont} 
\usepackage{times}
\usepackage{latexsym}
\usepackage{fontawesome}
\usepackage[linesnumbered]{algorithm2e}
\SetAlFnt{\footnotesize}
\SetAlCapFnt{\footnotesize}
\SetAlCapNameFnt{\footnotesize}
\SetKwComment{Comment}{/* }{ */}
\RestyleAlgo{ruled}
\usepackage[T1]{fontenc}
\usepackage[utf8]{inputenc}
\usepackage{inconsolata}
\usepackage{graphicx}
\usepackage{amsmath,amsfonts,amssymb}
\usepackage{fixltx2e}

\usepackage{xcolor}
\definecolor{ggreen}{HTML}{1F945D}
\definecolor{rred}{HTML}{D64531}
\usepackage{calc}
\newcommand{\fixedformat}[1]{\makebox[\widthof{00.00}][r]{{#1}}}
\newcommand{\fixedsubscript}[1]{\textsubscript{\fixedformat{{{#1}}}}}
\newcommand{\emptysubscript}{\textsubscript{\fixedformat{{{\phantom{0.00}}}}}}

\title{Better Benchmarking LLMs for Zero-Shot Dependency Parsing}
\author{Ana Ezquerro, Carlos Gómez-Rodríguez, David Vilares \\
  Universidade da Coruña, CITIC \\
  Departamento de Ciencias de la Computación y Tecnologías de la Información \\
  Campus de Elviña s/n, 15071 \\
  A Coruña, Spain\\
  \texttt{\{ana.ezquerro, carlos.gomez, david.vilares\}@udc.es}} 
\usepackage{multirow}
\newcommand{\best}[1]{\textbf{#1}}

\begin{document}
\maketitle
\begin{abstract}
While LLMs excel in zero-shot tasks, their performance in linguistic challenges like syntactic parsing has been less scrutinized. This paper studies
state-of-the-art open-weight LLMs on the task by comparing them to baselines that do not have access to the input sentence, including baselines that have not been used in this context such as random projective trees or optimal linear arrangements. The results show that most of the tested LLMs cannot outperform the best uninformed baselines, with only the newest and largest versions of LLaMA doing so for most languages, and still achieving rather low performance. Thus, accurate zero-shot syntactic parsing is not forthcoming with open LLMs.
\end{abstract}

\section{Introduction}
Autoregressive large language models (LLMs) and instruction-based variants \cite{jiang-2023-mistral7b, openai-2024-gpt4, llama3-2024} are known for their zero-shot and few-shot abilities \cite{radford-2019-language}. In practical terms, they can serve as versatile systems whose behavior is easily adapted through prompting. Beyond what we experience as everyday users, documented examples in the context of natural language processing (NLP) include question answering  \cite{baek-etal-2023-knowledge-augmented, li-etal-2024-self-prompting}, summarization \cite{wang-etal-2023-zero}, machine translation \cite{johnson-etal-2017-googles, wang-etal-2021-rethinking-zero, zhang-etal-2023-machine} and information retrieval \cite{ zhuang-etal-2023-open,adeyemi-etal-2024-zero, qin-etal-2024-large}, among many other tasks.

Related, syntactic parsing has long explored few-shot learning approaches. Prior to the development of current LLMs, various methods were studied to perform zero-shot or few-shot parsing, and many of these approaches achieved  competitive results. These methods focused on factors such as the quality and quantity of annotations \cite{meechan-maddon-nivre-2019-parse}, cross-lingual learning \cite{xu-koehn-2021-zero}, multilingual pre-training \cite{tran-bisazza-2019-zero}, and treebank difficulty \cite{sogaard-2020-languages, anderson-etal-2021-replicating}. However, the effectiveness of zero-shot parsing with LLMs remains a topic of debate. Recent work has showed how state-of-the-art LLMs exhibit low performance in syntactic parsing \cite{bai-2023-constituency, lin-2023-chatgpt}, even when designing manual specific prompts \cite{li-etal-2023-llm, blevins-etal-2023-prompting}. Nonetheless, these results have been deemed sufficient to categorize LLMs as potential zero-shot parsers. While some studies \cite{tian-etal-2024-large} suggest that multi-stage complex approaches can yield a competitive zero-shot performance, in this work we will focus on single prompt approaches -- similar to what works for other NLP tasks -- to evaluate to what extent LLMs can perform the task on their own without externally-provided planning.
Studies covering these approaches leave a substantial gap in evaluating LLMs on low-resource setups and often omit comparison with uninformed baselines, which are essential for determining whether LLMs achieve accuracy levels meaningfully above chance.

\paragraph{Contribution} We address the lack of comparison against uninformed baselines, and include some that have not been proposed before but offer a higher standard than traditional blind baselines, such as left- or right-branching trees. These baselines provide more robust benchmarks and offer a fairer evaluation of LLMs' potential as zero-shot parsers. We prioritize depth over breadth by evaluating a wide range of LLMs to identify any substantial differences across them - a contribution that, to our knowledge, has not been thoroughly explored in previous work.\footnote{Code available at \href{https://github.com/anaezquerro/naipar}{\texttt{github.com/anaezquerro/naipar}}}

\section{Zero-shot dependency parsing}
Next, we review the notation, benchmarks, uninformed baselines, and introduce the LLMs used. Dependency parsing is the task of obtaining the syntactic structure of a sentence as a set of labeled directed relations (\emph{dependencies}) between words. In zero-shot parsing (whether relying on LLMs or other models), the core idea is to perform the dependency parsing task without using task-specific labeled data during either the pre-training or fine-tuning steps. This approach contrasts with the standard setup for training dependency parsers, where task-specific labeled data is integral to explicitly teaching models syntactic structures in a supervised learning framework. Instead, zero-shot parsing leverages the model's general pre-trained knowledge to infer syntactic relationships in unseen data. Before the emergence of large autoregressive generative models, pre-training largely avoided task-specific annotations, aligning closely with the zero-shot paradigm. However, given the extensive and diverse nature of the data these models are trained on, it is plausible that some exposure to annotated dependency parsing examples has occurred. This possibility will be examined further in subsequent sections.

\paragraph{Notation} Let $W$=$(w_1,...,w_n)$ be a sentence, a dependency graph is defined as $G=(W,A)$, where $W$ is the set of nodes and $A$ the set of arcs. Each arc in $A$ is a tuple $(h,d)$, where $h\in[0,n]$ is the position of the head node, and $d \neq h \in[1,n]$ the position of the dependent node.\footnote{Graphs have arcs labeled with syntactic functions, but we ignore them here as our evaluation is unlabeled.} $G$ is a tree $T$ iff (i) is a \emph{connected acyclic} graph, (ii) each word $w_i$ has only one head, so $A=\{(h,d): d = 1...n\}$, and (iii) there is only one arc of the form \( (0, d) \), and \( w_d \) is designated as the \emph{root} of the sentence. This work only studies trees.

\subsection{Zero-shot (uninformed) baselines}
Previous work has reported results on parsing using LLMs \cite{lin-2023-chatgpt}, classifying LLMs as \emph{potentially} zero-shot parsers. We revisit this claim by proposing a comprehensive benchmark and comparing against uninformed baselines (i.e., baselines that generate an output tree without looking at the contents of the input, although sometimes with access to its length). Uninformed baselines are useful to determine whether the models are meaningfully processing the input or just generating outputs that could be obtained by chance or by using properties that are not specific to the input sentence (e.g. the common trend towards projectivity in human syntax). We use both conventional uninformed baselines that have been used previously in related contexts \citep[e.g.][]{klein-manning-2004-corpus} and more sophisticated, though still uninformed, baselines that, to our knowledge, have not yet been applied for this purpose.

\subsubsection{Conventional baselines}

We now describe baselines that have been used as naive approaches to build simplistic yet valid trees. 

\paragraph{Randomized root-based tree generation} Our most basic baseline randomly selects a root node, denoted as $d'$, and creates a dependency from $d'$ to the rest of nodes. Formally,  $\hat{A}=\{ (0, d'), (d', d) : d=(1,...,n) \neq d'\}$.

\paragraph{Right- and left-branching tree generation} This method assigns each word as a dependent of the previous (next) word, with the first (last) word as the root. Right-branching trees are a classic baseline for unsupervised English dependency parsing~\citep{klein-manning-2004-corpus}, as English syntax is predominantly right-branching. The left-branching baseline is included as some languages are predominantly left-branching.

\paragraph{Uniformly random tree generation} Another parsing baseline~\citep{klein-manning-2004-corpus}. We use the \citet{aldous-1990-random} algorithm to guarantee generation of a uniformly random dependency tree.

\paragraph{Sampling from a reference treebank} We build the tree distributions of different lengths from a reference treebank. For a sentence, we sampled a dependency tree from the distribution of length $n$. Note that this is the only one among our baselines that has access to a treebank, although it \emph{is still uninformed} with respect to the input sentence.

\subsubsection{Novel uninformed baselines}
We refine random tree generation by taking into account observed properties of human language: the scarcity of crossing dependencies~\citep{FerGomEstPhysA2018} and dependency distance minimization, i.e., the tendency of syntactic structures to minimize the distance between syntactically related words (i.e. the length of dependencies) in order to reduce cognitive processing effort~\citep{liu2017dependency,ferreri2020optimality}.

\paragraph{Uniformly random projective tree generation} The goal is to generate a projective tree  (where dependencies do not cross) uniformly at random. As rejection sampling is too slow, we use \citet{Nijenhuis1978}'s algorithm to generate a random unlabeled rooted tree and then assign a random projective arrangement following \citep{Futrell2015a,projarr}.\footnotemark[3]

\paragraph{Uniformly random (projective) optimal-distance tree generation} Again, we start from a uniformly random unlabeled rooted tree. In this case, we give it the linear arrangement that minimizes the sum of dependency distances, using ~\citet{Shiloach1979}'s algorithm, as well as the minimum-distance \emph{projective} arrangement, with the algorithm by~\citet{ALEMANYPUIG2022106204}.\footnotemark[3]
\footnotetext[3]{We used the implementation of these algorithms in the LAL library~\citep{alemany-puig-etal-2021-linear}.}

\subsection{Zero-shot parsing with LLMs}
\paragraph{Prompting setup} Adopting a strategy similar to \citet{lin-2023-chatgpt}, we query LLMs using simple prompts. The prompt includes an introductory sentence requesting output in CoNLL format, followed by a basic example from a reference treebank, where only the \texttt{ID}, \texttt{HEAD}, and \texttt{DEPREL} fields are populated. We selected a random sentence of length \(4\) to \(7\) to avoid longer sequences, maintaining a zero-shot setup. Although this may resemble a one-shot setup, the example is intentionally simple, serving only to reduce formatting errors rather than offering linguistic content. Figure \ref{fig:prompt-example} breaks down the specific prompt we used.

\paragraph{Postprocessing} We account for possible corrupted outputs, such as column mismatches, missing nodes, or multiple roots. From the model's raw output, we applied two post-processing steps: first, filtering tabular lines and filling fields to match the CoNLL format with correct row and column counts for sentences of length \( n \). Second, we resolved cycles, enforced a unique-root constraint, and replaced out-of-range arcs with root connections to ensure a single-rooted, connected tree. Figure \ref{fig:prompt-example} also shows an example of the input and output after the first post-processing step. 

\begin{figure}[tbp]
    \centering\scriptsize
    \begin{tabular}{p{7.2cm}}
        \hline 
        \textbf{Prompt example} \\ 
        \hline 
        \texttt{In dependency parsing the CoNLL format for the sentence <The trial begins again Nov 28 .> is:}\\
        \texttt{1	The	\_	\_	\_	\_	2	det	\_	\_} \\ 
        \texttt{2	trial	\_	\_	\_   \_	3	nsubj	\_	\_} \\ 
        \texttt{3	begins	\_	\_	\_	\_	0	root	\_	\_} \\
        \texttt{4	again	\_	\_	\_	\_	3	advmod	\_	\_} \\ 
        \texttt{5	Nov.	\_	\_	\_	\_	3	obl:tmod	\_	\_}\\ 
        \texttt{6	28	\_	\_	\_	\_	5	nummod	\_	\_}\\ 
        \texttt{7	.	\_	\_	\_	\_	3	punct	\_	\_} \\
        \texttt{Now return the CoNLL format for the sentence: <What if Google Morphed Into GoogleOS ?>} \\
        \hline \textbf{(1) Well-formatted output }\\ \hline 
        \texttt{1	What	\_	\_	\_	\_	0	nsubj	\_	\_} \\ 
        \texttt{2	if	\_	\_	\_	\_	4	mark	\_	\_ \_} \\ 
        \texttt{3 Google	\_	\_	\_	\_	4	nsubj	\_	\_} \\ 
        \texttt{4	Morphed	\_	\_	\_	\_	0	root	\_	\_} \\ 
        \texttt{5	into	\_	\_	\_	\_	6	case	\_	\_} \\ 
        \texttt{6	GoogleOS	\_	\_	\_	\_	8	nmod	\_	\_} \\ 
        \texttt{7	?	\_	\_	\_	\_	4	punct	\_	\_}\\
        \hline 
    \end{tabular}
    \caption{\label{fig:prompt-example} Prompt and output after the first post-processing. See Figure  \ref{fig:prompt-example-complete} for step-by-step process. }
\end{figure}

\begin{table*}[thpb!]
    \setlength{\tabcolsep}{1pt}
    \centering\footnotesize
    \begin{tabular}{|p{0.3cm}p{1.5cm}|rrr|rrr|rrr|rrr|}
        \cline{3-14} 
        \multicolumn{2}{c|}{}& \multicolumn{3}{c|}{\bf English\,\textsubscript{EWT}} & \multicolumn{3}{c|}{\bf French\,\textsubscript{GSD}} & \multicolumn{3}{c|}{\bf German\,\textsubscript{GSD}} & \multicolumn{3}{c|}{\bf Hindi\,\textsubscript{HDTB}}\\
        \cline{3-14}
        \multicolumn{2}{c|}{}& \multicolumn{1}{l}{\textbf{UAS}} & \multicolumn{1}{l}{\textbf{UM}} & \multicolumn{1}{l|}{\%\textbf{w}} & \multicolumn{1}{l}{\textbf{UAS}} & \multicolumn{1}{l}{\textbf{UM}} & \multicolumn{1}{l|}{\%\textbf{w}} & \multicolumn{1}{l}{\textbf{UAS}} & \multicolumn{1}{l}{\textbf{UM}} & \multicolumn{1}{l|}{\%\textbf{w}} & \multicolumn{1}{l}{\textbf{UAS}} & \multicolumn{1}{l}{\textbf{UM}} & \multicolumn{1}{l|}{\%\textbf{w}} \\ 
        \hline 
        & A    & 20.74\emptysubscript & 13.91\emptysubscript & 100.00 & 5.99\emptysubscript & 0.24\emptysubscript & 100.00 & 9.79\emptysubscript & 2.05\emptysubscript & 100.00 & 5.92\emptysubscript & 0.06\emptysubscript & 100.00 \\
        & R    & 23.30\emptysubscript & 12.13\emptysubscript & 100.00 & 10.67\emptysubscript & 0.00\emptysubscript & 100.00 & 11.59\emptysubscript & 1.23\emptysubscript & 100.00 & \best{25.34}\emptysubscript & 0.00\emptysubscript & 100.00 \\
        & L    & 34.41\emptysubscript & 9.39\emptysubscript & 100.00 & 29.78\emptysubscript & 0.00\emptysubscript & 100.00 & 29.59\emptysubscript & 1.43\emptysubscript & 100.00 & 21.34\emptysubscript & 0.00\emptysubscript & 100.00 \\
        & RD   & 20.10\emptysubscript & 10.78\emptysubscript & 100.00 & 5.93\emptysubscript & 0.00\emptysubscript & 100.00 & 9.64\emptysubscript & 1.13\emptysubscript & 100.00 & 5.67\emptysubscript & 0.00\emptysubscript & 100.00 \\
        & RD*  & 21.45\emptysubscript & 12.42\emptysubscript & 100.00 & 6.06\emptysubscript & 0.00\emptysubscript & 100.00 & 9.38\emptysubscript & 1.74\emptysubscript & 100.00 & 5.59\emptysubscript & 0.00\emptysubscript & 100.00 \\
        & LI   & 28.09\emptysubscript & 11.75\emptysubscript & 100.00 & 16.81\emptysubscript & 0.00\emptysubscript & 100.00 & 19.24\emptysubscript & 1.64\emptysubscript & 100.00 & 19.99\emptysubscript & 0.06\emptysubscript & 100.00 \\
        & LI*  & 26.99\emptysubscript & 10.98\emptysubscript & 100.00 & 17.08\emptysubscript & 0.00\emptysubscript & 100.00 & 20.53\emptysubscript & 1.84\emptysubscript & 100.00 & 20.06\emptysubscript & 0.00\emptysubscript & 100.00 \\
        & S    & 31.14\emptysubscript & 15.55\emptysubscript & 100.00 & 19.43\emptysubscript & \best{0.96}\emptysubscript & 100.00 & 20.02\emptysubscript & 2.46\emptysubscript & 100.00 & 17.92\emptysubscript & \best{0.30}\emptysubscript & 100.00 \\
        \hline
        \multirow{3}{*}{\raisebox{-0.1\height}{\includegraphics[height=0.3cm]{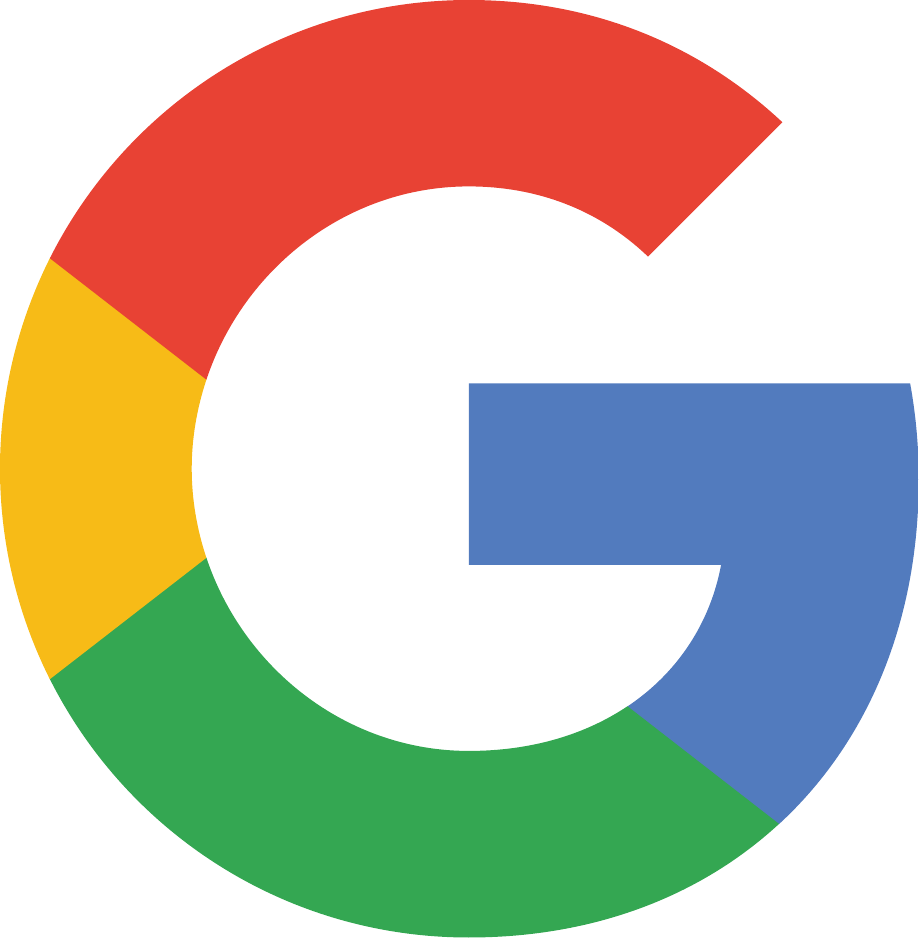}}} 
        & \texttt{v1-2b}    & 15.80\textcolor{ggreen}{\fixedsubscript{5.23}} & 5.54\textcolor{ggreen}{\fixedsubscript{7.22}} & 7.17 & 6.51\textcolor{ggreen}{\fixedsubscript{0.52}} & 0.00\textcolor{ggreen}{\fixedsubscript{0.00}} & 0.48 & 9.66\textcolor{ggreen}{\fixedsubscript{0.91}} & 1.02\textcolor{ggreen}{\fixedsubscript{0.92}} & 2.15 & 11.96\textcolor{rred}{ \fixedsubscript{-1.03}} & 0.00\textcolor{ggreen}{\fixedsubscript{0.00}} & 6.92 \\
        & \texttt{v1-7b}    &   21.26\textcolor{ggreen}{\fixedsubscript{5.17}} & 6.93\textcolor{ggreen}{\fixedsubscript{6.88}} & 24.84 & 14.93\textcolor{rred}{ \fixedsubscript{-0.66}} & 0.00\textcolor{ggreen}{\fixedsubscript{0.00}} & 7.69 & 16.61\textcolor{ggreen}{\fixedsubscript{0.99}} & 0.82\textcolor{ggreen}{\fixedsubscript{1.13}} & 10.75 & 9.78\textcolor{rred}{ \fixedsubscript{-0.08}} & 0.00\textcolor{ggreen}{\fixedsubscript{0.00}} & 8.43 \\
        & \texttt{v2-9b}    &  20.20\textcolor{ggreen}{\fixedsubscript{3.35}} & 6.79\textcolor{ggreen}{\fixedsubscript{6.50}} & 15.17 & 13.32\textcolor{rred}{ \fixedsubscript{-1.34}} & 0.00\textcolor{ggreen}{\fixedsubscript{0.72}} & 4.81 & 14.75\textcolor{rred}{ \fixedsubscript{-0.68}} & 0.92\textcolor{ggreen}{\fixedsubscript{1.13}} & 5.32 & 12.18\textcolor{rred}{ \fixedsubscript{-0.77}} & 0.00\textcolor{ggreen}{\fixedsubscript{0.06}} & 3.27 \\
        \hline 
        \multirow{9}{*}{\includegraphics[height=0.22cm]{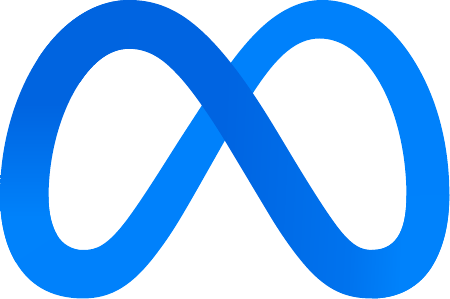}} 
        & \texttt{v2-7b}    & 12.98\textcolor{ggreen}{\fixedsubscript{10.38}} & 3.18\textcolor{ggreen}{\fixedsubscript{11.12}} & 24.22 & 18.20\textcolor{rred}{ \fixedsubscript{-1.45}} & 0.00\textcolor{ggreen}{\fixedsubscript{0.24}} & 1.92 & 18.70\textcolor{ggreen}{\fixedsubscript{0.19}} & 0.10\textcolor{ggreen}{\fixedsubscript{2.05}} & 4.09 & 10.64\textcolor{rred}{ \fixedsubscript{-1.74}} & 0.00\textcolor{ggreen}{\fixedsubscript{0.00}} & 5.82 \\
        & \texttt{v2-13b}   & 18.95\textcolor{ggreen}{\fixedsubscript{4.01}} & 5.83\textcolor{ggreen}{\fixedsubscript{8.96}} & 14.59 & 13.64\textcolor{rred}{ \fixedsubscript{-0.81}} & 0.00\textcolor{ggreen}{\fixedsubscript{0.00}} & 22.12 & 19.40\textcolor{rred}{ \fixedsubscript{-1.78}} & 0.00\textcolor{ggreen}{\fixedsubscript{0.00}} & 32.89 & 14.77\textcolor{rred}{ \fixedsubscript{-1.77}} & 0.00\textcolor{ggreen}{\fixedsubscript{0.06}} & 18.71 \\
        & \texttt{v2-70b}   &   13.78\textcolor{ggreen}{\fixedsubscript{11.38}} & 4.00\textcolor{ggreen}{\fixedsubscript{10.98}} & 46.89 & 19.05\textcolor{rred}{ \fixedsubscript{-1.42}} & 0.24\textcolor{ggreen}{\fixedsubscript{0.00}} & 13.70 & 25.88\textcolor{rred}{ \fixedsubscript{-2.41}} & 1.43\textcolor{ggreen}{\fixedsubscript{0.20}} & 23.34 & 15.27\textcolor{rred}{ \fixedsubscript{-2.33}} & 0.00\textcolor{ggreen}{\fixedsubscript{0.00}} & 7.47 \\
        & \texttt{v3-8b}    &  18.34\textcolor{ggreen}{\fixedsubscript{6.56}} & 7.41\textcolor{ggreen}{\fixedsubscript{8.09}} & 49.01 & 14.80\textcolor{rred}{ \fixedsubscript{-2.84}} & 0.00\textcolor{ggreen}{\fixedsubscript{0.00}} & 4.81 & 29.51\textcolor{rred}{ \fixedsubscript{-2.26}} & 1.74\textcolor{ggreen}{\fixedsubscript{0.10}} & 22.42 & 17.03\textcolor{rred}{ \fixedsubscript{-1.54}} & 0.00\textcolor{ggreen}{\fixedsubscript{0.06}} & 0.48 \\
        & \texttt{v3-70b}   &  38.30\textcolor{ggreen}{\fixedsubscript{0.98}} & \best{16.37}\textcolor{ggreen}{\fixedsubscript{1.44}} & 58.69 & 29.20\textcolor{rred}{ \fixedsubscript{-0.61}} & 0.96\textcolor{ggreen}{\fixedsubscript{0.48}} & 33.41 & 33.91\textcolor{rred}{ \fixedsubscript{-1.16}} & 3.17\textcolor{ggreen}{\fixedsubscript{0.41}} & 28.25 & 14.24\textcolor{rred}{ \fixedsubscript{-1.54}} & 0.00\textcolor{ggreen}{\fixedsubscript{0.00}} & 26.14 \\
        & \texttt{v3.1-8b}  &  28.83\textcolor{rred}{ \fixedsubscript{-1.31}} & 11.75\textcolor{ggreen}{\fixedsubscript{2.07}} & 34.38 & 24.61\textcolor{rred}{ \fixedsubscript{-6.35}} & 0.72\textcolor{ggreen}{\fixedsubscript{0.00}} & 5.29 & 26.93\textcolor{rred}{ \fixedsubscript{-4.67}} & 1.64\textcolor{ggreen}{\fixedsubscript{0.20}} & 12.90 & 18.86\textcolor{rred}{ \fixedsubscript{-2.11}} & 0.00\textcolor{ggreen}{\fixedsubscript{0.12}} & 2.43 \\
        & \texttt{v3.1-70b} &  \best{39.69}\textcolor{ggreen}{\fixedsubscript{1.46}} & 15.65\textcolor{ggreen}{\fixedsubscript{2.41}} & \best{65.86} & \best{34.62}\textcolor{rred}{ \fixedsubscript{-1.30}} & 0.96\textcolor{ggreen}{\fixedsubscript{0.24}} & \best{42.79} & \best{36.75}\textcolor{rred}{ \fixedsubscript{-1.14}} & \best{3.48}\textcolor{ggreen}{\fixedsubscript{0.20}} & \best{46.57} & 14.37\textcolor{rred}{ \fixedsubscript{-0.69}} & 0.00\textcolor{ggreen}{\fixedsubscript{0.00}} & 26.14 \\
        & \texttt{v3.2-1b}&  15.07\textcolor{ggreen}{\fixedsubscript{7.35}} & 4.53\textcolor{ggreen}{\fixedsubscript{8.33}} & 16.75 & 8.05\textcolor{rred}{\fixedsubscript{-0.65}} & 0.00\textcolor{ggreen}{\fixedsubscript{0.24}} & 11.06 & 12.61\textcolor{rred}{\fixedsubscript{-0.9}} & 0.41\textcolor{ggreen}{\fixedsubscript{1.64}} & 10.03 & 7.20\textcolor{rred}{\fixedsubscript{-1.9}} & 0.00\textcolor{ggreen}{\fixedsubscript{0.0}} & 8.37   \\
        & \texttt{v3.2-3b} &  18.51\textcolor{ggreen}{\fixedsubscript{3.98}} & 6.64\textcolor{ggreen}{\fixedsubscript{6.55}} & 18.68 & 10.22\textcolor{rred}{\fixedsubscript{-1.1}} & 0.24\textcolor{ggreen}{\fixedsubscript{0.0}} & 12.74 & 17.69\textcolor{ggreen}{\fixedsubscript{0.34}} & 1.13\textcolor{ggreen}{\fixedsubscript{1.13}} & 10.44 & 13.84\textcolor{ggreen}{\fixedsubscript{0.03}} & 0.00\textcolor{ggreen}{\fixedsubscript{0.0}} & 14.90  \\
        \hline 
        \multirow{8}{*}{\raisebox{-0.1\height}{\includegraphics[height=0.3cm]{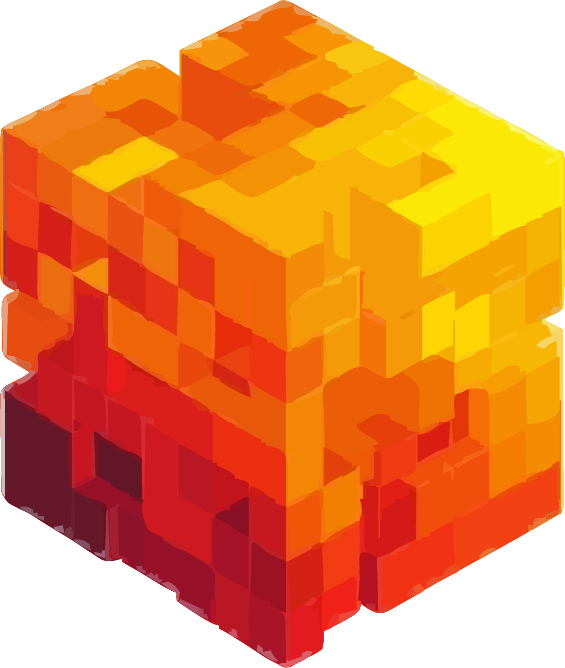}}} 
        & \texttt{v1-7b}    &  18.59\textcolor{ggreen}{\fixedsubscript{3.83}} & 6.55\textcolor{ggreen}{\fixedsubscript{6.55}} & 16.85 & 10.32\textcolor{rred}{ \fixedsubscript{-0.54}} & 0.00\textcolor{ggreen}{\fixedsubscript{0.00}} & 5.29 & 16.37\textcolor{rred}{ \fixedsubscript{-1.31}} & 1.23\textcolor{ggreen}{\fixedsubscript{0.61}} & 5.94 & 10.63\textcolor{rred}{ \fixedsubscript{-0.44}} & 0.00\textcolor{ggreen}{\fixedsubscript{0.00}} & 0.12 \\
        & \texttt{v2-7b}    & 23.02\textcolor{ggreen}{\fixedsubscript{2.10}} & 6.93\textcolor{ggreen}{\fixedsubscript{6.07}} & 15.12 & 18.73\textcolor{rred}{ \fixedsubscript{-3.28}} & 0.24\textcolor{ggreen}{\fixedsubscript{0.00}} & 3.85 & 20.49\textcolor{rred}{ \fixedsubscript{-2.90}} & 1.13\textcolor{ggreen}{\fixedsubscript{0.61}} & 5.02 & 13.80\textcolor{rred}{ \fixedsubscript{-1.11}} & 0.00\textcolor{ggreen}{\fixedsubscript{0.06}} & 0.36 \\
        & \texttt{v3-7b}    & 25.04\textcolor{ggreen}{\fixedsubscript{2.49}} & 7.66\textcolor{ggreen}{\fixedsubscript{5.39}} & 28.17 & 27.36\textcolor{rred}{ \fixedsubscript{-5.52}} & 0.24\textcolor{ggreen}{\fixedsubscript{0.48}} & 11.06 & 28.34\textcolor{rred}{ \fixedsubscript{-4.98}} & 1.13\textcolor{ggreen}{\fixedsubscript{0.51}} & 19.55 & 19.41\textcolor{rred}{ \fixedsubscript{-3.15}} & 0.00\textcolor{ggreen}{\fixedsubscript{0.00}} & 0.71 \\
        & \texttt{x1-7b}    &  15.46\textcolor{ggreen}{\fixedsubscript{3.21}} & 2.63\textcolor{ggreen}{\fixedsubscript{4.17}} & 26.22 & 13.00\textcolor{rred}{ \fixedsubscript{-1.06}} & 0.24\textcolor{ggreen}{\fixedsubscript{0.00}} & 3.37 & 16.67\textcolor{ggreen}{\fixedsubscript{0.00}} & 0.00\textcolor{ggreen}{\fixedsubscript{0.00}} & 25.00 & 13.68\textcolor{rred}{ \fixedsubscript{-0.67}} & 0.06\textcolor{ggreen}{\fixedsubscript{0.00}} & 1.25 \\
        & \texttt{x1-22b} &32.91\textcolor{ggreen}{\fixedsubscript{0.85}} & 13.72\textcolor{ggreen}{\fixedsubscript{3.32}} & 57.74 & 23.75\textcolor{ggreen}{\fixedsubscript{0.13}} & 0.68\textcolor{ggreen}{\fixedsubscript{0.24}} & 36.73 & 22.48\textcolor{rred}{\fixedsubscript{-0.19}} & 2.76\textcolor{ggreen}{\fixedsubscript{0.92}} & 38.44 & 19.37\textcolor{rred}{\fixedsubscript{-0.67}} & 0.09\textcolor{ggreen}{\fixedsubscript{0.26}} & \best{37.84}   \\
        & \texttt{nemo}     & 20.96\textcolor{ggreen}{\fixedsubscript{3.89}} & 7.56\textcolor{ggreen}{\fixedsubscript{7.03}} & 15.74 & 15.85\textcolor{rred}{ \fixedsubscript{-1.49}} & 0.00\textcolor{ggreen}{\fixedsubscript{0.00}} & 3.61 & 14.10\textcolor{rred}{ \fixedsubscript{-0.57}} & 1.23\textcolor{ggreen}{\fixedsubscript{0.72}} & 4.09 & 9.59\textcolor{rred}{ \fixedsubscript{-0.06}} & 0.00\textcolor{ggreen}{\fixedsubscript{0.00}} & 0.48 \\
        & \texttt{large}    &  28.71\textcolor{ggreen}{\fixedsubscript{0.81}} & 10.01\textcolor{ggreen}{\fixedsubscript{4.77}} & 18.25 & 15.21\textcolor{rred}{ \fixedsubscript{-0.58}} & 0.83\textcolor{ggreen}{\fixedsubscript{0.42}} & 5.00 & 17.66\textcolor{ggreen}{\fixedsubscript{0.58}} & 1.74\textcolor{ggreen}{\fixedsubscript{0.92}} & 7.88 & 14.46\textcolor{rred}{ \fixedsubscript{-1.4}} & 0.00\textcolor{ggreen}{\fixedsubscript{0.00}} & 26.14 \\
    \hline 
    \end{tabular}
    \caption{\label{tab:results}Performance on the test sets. The baselines are: all-to-root (A), left (L) and right (R) branching, random generation (RD), optimal linear arrangement (LI) and sampling (S). The symbol (*) indicates if projectivity is fixed as a constraint. \%\textbf{w} is the ratio of outputs that did not require post-processing.  
    We also report results with Gemma (\raisebox{-0.1\dimexpr\ht\strutbox}{\includegraphics[height=0.3cm]{images/google.pdf}}), LlaMA (\includegraphics[height=0.22cm]{images/meta.pdf}), and Mistral models (\raisebox{-0.1\dimexpr\ht\strutbox}{\includegraphics[height=0.3cm]{images/mistral.pdf}}), with versions (\texttt{v}, \texttt{x}) and parameter counts. Subscripts indicate performance boost from the second post-processing step.}
\end{table*}

\section{Experiments}\label{section-experiments}
We conduct an in-depth evaluation of LLMs as zero-shot dependency parsers by generating outputs in CoNLL format and comparing them to uninformed baselines. Unlike \citet{lin-2023-chatgpt}, who evaluated only ChatGPT-3.5 due to limited system availability, our work expands the analysis to a broader set of models across a select few languages, albeit on a smaller subset of treebanks. This approach, while time-intensive due to the extensive input and output token requirements, offers a more comprehensive understanding of model performance across different LLMs.

\paragraph{Datasets} We selected 4 treebanks from UD 2.14 \cite{ud214} to conduct experiments in different languages, specifically in English\textsubscript{EWT}, French\textsubscript{GSD}, German\textsubscript{GSD}, and Hindi\textsubscript{HDTB}. 

\paragraph{Evaluation} We use the unlabeled attachment score (UAS) and unlabeled exact match (UM) as our primary metrics. For the zero-shot dependency parsers, we report performance after the first post-processing step (ensuring that the CoNLL format file contains all columns) and the second (confirming that the tree is well-formed).

\paragraph{Models} We selected several instruction-based models from the Gemma \cite{gemma-2024, gemma2-2024}, LLaMA \cite{llama2-2023, llama3-2024}, and Mistral \cite{jiang-2023-mistral7b, jiang-2024-mixtral} series. Appendix \ref{ap:experiments} (Table \ref{tab:hf-models}) breaks down the links to all models. All reported results were obtained limiting the inference  to half precision.\footnote{Preliminary experiments indicated that reduced inference precision had minimal impact on performance.}

\section{Analysis of results}
Table \ref{tab:results} compares the performance of the tested models with uninformed baselines. We see that only the latest and largest versions of LLaMa (i.e., the 70B versions of Llama 3 and 3.1) consistently outperform all the baselines in most languages in terms of UAS and UM, and only do so barely (e.g., with the best result on English being about 5.5 points above the left-branching baseline without postprocessing, and close to 7 points with postprocessing). The rest of the models clearly fall behind, showing that they are not doing any meaningful parsing at all. In the case of Hindi, no model at all reaches the best baselines. Among our baselines, traditional left (or right in the case of Hindi) branching baselines are the most competitive in terms of UAS,\footnote{Superiority of the left-branching baseline on English can be surprising, as right-branching has often been deemed better on English unsupervised parsing~\citep{klein-manning-2004-corpus,li-etal-2020-empirical}; but these papers 
do not use UD.} although baselines based on optimal linear arrangement come close, and the sampling baseline is better in terms of English UM. In Appendix~\ref{ap:experiments} (Tables \ref{tab:en-pos} to \ref{tab:hi-pos}) we also include tables showing the individual scores of each model based on the PoS tag of the head in each treebank.

\begin{figure}[htbp!]
    \centering
    \includegraphics[width=\linewidth]{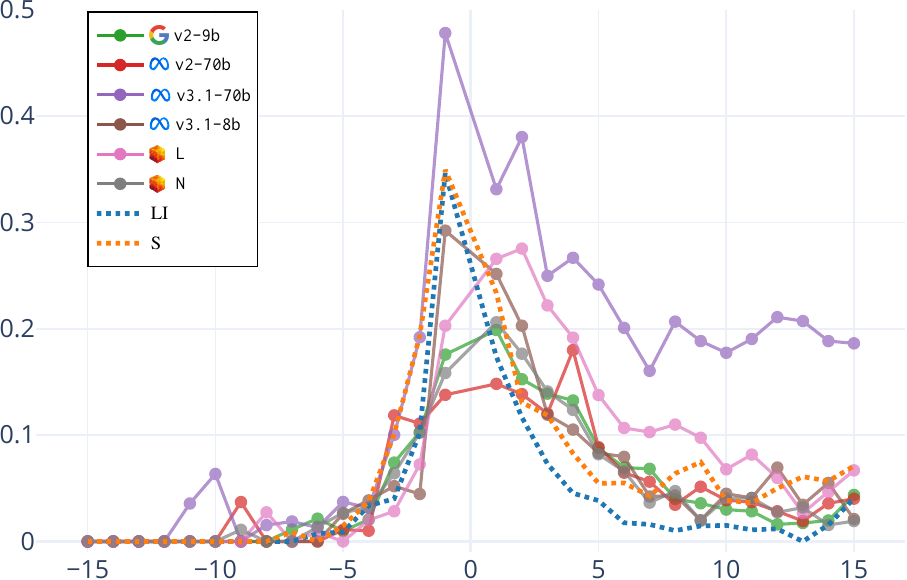}
    \caption{\label{fig:displ}F-score across displacements 
    in the English\textsubscript{EWT} test set.}
\end{figure}

Figure \ref{fig:displ} complements Table \ref{tab:results} by illustrating the performance of a representative subset of models in terms of dependency displacements (i.e., performance taking into account the difference between the position of the dependent and its head) for the English\textsubscript{EWT} treebank. We observe that LLaMa v3.1 70B consistently performs better than the sampling and optimal linear arrangement baselines, not only on short dependencies but also on longer rightward dependencies. However, for the rest of the models, the differences with respect to uninformed baselines become subtler. Similar figures for the other evaluated treebanks can be found in the Appendix (Figures \ref{fig:fr-displ}, \ref{fig:de-displ}, and \ref{fig:hi-displ}).

Overall, the results show that open-weight LLMs are far from being potential zero-shot dependency parsers, contrary to claims about ChatGPT~\cite{lin-2023-chatgpt}. Considerable scaling or other improvements would be required for this situation to change.

\section{Limitations}

\paragraph{Memorization}  
Memorization refers to the LLM's ability to recall specific patterns, structures, or dependencies encountered during pre-training, rather than generalizing to unseen cases \cite{hartmann2023sok}. This poses a risk of generalization issues or the regurgitation of chunks of text, which could affect our evaluation but is difficult to quantify~\citep{sainz-etal-2023-nlp}. Although this is beyond the main scope of our work, we have attempted to briefly analyze this phenomenon. To do so, we crawled a few hundred recent news articles from the New York Times Archive API.\footnote{\href{https://developer.nytimes.com/docs/archive-product/1}{\texttt{developer.nytimes.com/docs/archive-product/1}}} The aim was to collect new text, guaranteeing that no annotations for it were available online when the models were trained. We then produced silver annotations by using a trained  graph-based model \cite{dozat-etal-2017-stanfords} -- a state-of-the-art dependency parser -- to parse these articles. In Table \ref{tab:silver-results}, we present the results of a few representative models against these silver annotations. The results are consistent with those for UD in Table~\ref{tab:results}: while UAS scores are lower across the board, this happens both for LLMs and baselines, and likely stems from NYT sentences being longer on average. In relative terms, the same trends as in UD stand, with only LLaMa 3.1-70B clearly outperforming all baselines, so we do not detect evidence of our main results being overestimated due to data contamination.

\begin{table}[h]
    \centering\footnotesize
    \setlength{\tabcolsep}{5pt}
    \begin{tabular}{|p{0.3cm}p{1.5cm}|rrr|}
        \cline{3-5} 
        \multicolumn{2}{c|}{}& \multicolumn{1}{c}{\textbf{UAS}} & \multicolumn{1}{c}{\textbf{UM}} & \multicolumn{1}{c|}{\%\textbf{w}} \\ 
        \hline 
        & A & 6.46\textsubscript{\textcolor{rred}{-14.27}}	& 1.69\textsubscript{\textcolor{rred}{-12.22}} & 100 \emptysubscript \\
        & R & 24.31\fixedsubscript{\textcolor{ggreen}{1.01}}& 0.70\textsubscript{\textcolor{rred}{-11.43}} & 100 \emptysubscript\\
        & L & 20.15\textsubscript{\textcolor{rred}{-14.25}}& 2.32\fixedsubscript{\textcolor{rred}{-7.07}} & 100 \emptysubscript\\
        & RD & 5.51\textsubscript{\textcolor{rred}{-14.59}}& 0.91\textsubscript{\textcolor{rred}{-11.51}} & 100 \emptysubscript\\
        & RD*& 7.32\textsubscript{\textcolor{rred}{-14.13}}& 1.51\textsubscript{\textcolor{rred}{-10.91}} & 100 \emptysubscript\\
        & LI& 16.39\textsubscript{\textcolor{rred}{-11.70}}& 1.77\fixedsubscript{\textcolor{rred}{-9.98}} & 100 \emptysubscript\\
        & LI*& 19.54\fixedsubscript{\textcolor{rred}{-7.45}} & 1.40\fixedsubscript{\textcolor{rred}{-9.58}}& 100 \emptysubscript\\
        & S & 22.24\fixedsubscript{\textcolor{rred}{-8.89}}& 2.55\textsubscript{\textcolor{rred}{-13.00}} & 100 \emptysubscript\\
        \hline 
        \multirow{2}{*}{\includegraphics[height=0.22cm]{images/meta.pdf}} 
        & \texttt{v3.1-70b}    & 28.83\textcolor{rred}{\textsubscript{-12.32}} & 0.00\textcolor{rred}{\textsubscript{-18.06}} & 38.72\textsubscript{\textcolor{rred}{-27.14}} \\
        & \texttt{v3.2-3b}    & 8.14\textcolor{rred}{\textsubscript{-14.35}} & 0.00\textcolor{rred}{\textsubscript{-13.19}} & 11.04\fixedsubscript{\textcolor{rred}{-7.64}}\\
        \hline 
        \multirow{2}{*}{\raisebox{-0.1\height}{\includegraphics[height=0.3cm]{images/mistral.pdf}}} 
        & \texttt{x1-22b} &  20.37\textcolor{rred}{\textsubscript{-13.39}} & 1.00\textcolor{rred}{\textsubscript{-16.04}} & 58.92\fixedsubscript{\textcolor{ggreen}{1.18}} \\
        & \texttt{large} & 25.37\textcolor{rred}{\fixedsubscript{-4.15}} & 3.00\textcolor{rred}{\textsubscript{-11.78}} & 13.24\textsubscript{\textcolor{rred}{-5.01}} \\ 
        \hline 
    \end{tabular}
    \caption{\label{tab:silver-results}Performance on silver annotations. Subscripts denote the performance drop from Table \ref{tab:results}.}
\end{table}

\paragraph{Prompting} The prompting approach used in this study followed a straightforward design. We acknowledge that there may be room for improving parsing performance through more advanced prompt engineering techniques. Our goal was methodological, establishing a set of uninformed baselines rather than optimizing prompt configurations. In this context, approaches such as in-context learning \cite{brown-etal-2020-language,chen2021meta}, chain-of-thought prompting \cite{wei2022chain}, and self-consistency \cite{wang2022self} have shown promise in improving performance by fostering more structured reasoning. 

\paragraph{Language selection} Our selection of languages was limited to a small set, three of which belong to the Indo-European family. This choice was driven by two key factors. First, although we had the exclusive access to a few 24GB RTX 3090, these were insufficient for running larger models effectively. We also had access to CESGA, the supercomputing center of Galicia; but it was limited to queuing systems, making it difficult to estimate running times and prioritize experiments given the large number of models involved. Additionally, although many models claim to be multilingual, their performance tends to be skewed toward a subset of widely spoken languages. We therefore selected languages that have the most support across models to ensure consistent evaluations.\footnote{Note that not all languages are supported by all models (Table \ref{tab:support_grouped_corrected}). Our selection aims to include widely supported languages to ensure fair comparisons across models.}

\section{Conclusion}
We revisited the potential of autoregressive LLMs as zero-shot dependency parsers. Taking a more conservative approach than previous studies, we compared several LLMs with simple baselines to establish minimal performance benchmarks. Our results show that most LLMs performed on par with uninformed baselines, indicating comparable performance to toy approaches that operate without any access to the input sentence.

\section*{Acknowledgments}

We acknowledge grants SCANNER-UDC (PID2020-113230RB-C21) funded by MICIU/AEI/10.13039/501100011033; GAP (PID2022-139308OA-I00) funded by MICIU/AEI/10.13039/501100011033/ and ERDF, EU; LATCHING (PID2023-147129OB-C21) funded by MICIU/AEI/10.13039/501100011033 and ERDF, EU; and TSI-100925-2023-1 funded by Ministry for Digital Transformation and Civil Service and ``NextGenerationEU'' PRTR; as well as funding by Xunta de Galicia (ED431C 2024/02), and Centro de Investigación de Galicia ``CITIC'', funded by the Xunta de Galicia through the collaboration agreement between the Consellería de Cultura, Educación, Formación Profesional e Universidades and the Galician universities for the reinforcement of the research centres of the Galician University System (CIGUS). We also extend our gratitude to CESGA, the supercomputing center of Galicia, for granting us access to its resources.

\bibliographystyle{acl_natbib}
\bibliography{custom}
\appendix
\section{Post-processing} \label{ap:postprocessing}
Figure \ref{fig:prompt-example-complete} breaks down the process of obtaining a valid dependency tree (fully connected, no cycles and only one root) from the raw output of the LLMs. In the second post-processing step, to enforce the unique-root constraint we randomly selected a root from the subset of outgoing arcs from node \texttt{0}, or just a random node if there are no such arcs. Out-of-range arcs were resolved by replacing the head with the root node. To break cycles and connect all components, our post-processing algorithm performs a breadth-first search from the root node, removing those arcs that create cycles and connecting disconnected nodes to the root node.

\begin{figure}[h]
    \centering\scriptsize
    \begin{tabular}{p{7.2cm}}
        \hline 
        \textbf{Prompt example} \\ 
        \hline 
        \texttt{In dependency parsing the CoNLL format for the sentence <The trial begins again Nov 28 .> is:}\\
        \texttt{1	The	\_	\_	\_	\_	2	det	\_	\_} \\ 
        \texttt{2	trial	\_	\_	\_   \_	3	nsubj	\_	\_} \\ 
        \texttt{3	begins	\_	\_	\_	\_	0	root	\_	\_} \\
        \texttt{4	again	\_	\_	\_	\_	3	advmod	\_	\_} \\ 
        \texttt{5	Nov.	\_	\_	\_	\_	3	obl:tmod	\_	\_}\\ 
        \texttt{6	28	\_	\_	\_	\_	5	nummod	\_	\_}\\ 
        \texttt{7	.	\_	\_	\_	\_	3	punct	\_	\_} \\
        \texttt{Now return the CoNLL format for the sentence: <What if Google Morphed Into GoogleOS ?>} \\
        \hline \textbf{Raw output} \\ \hline 
        \texttt{Sure! This is the CoNLL format for the sentence <What if Google Morphed Into GoogleOS ?>} \\
        \texttt{1	What	\_	\_	\_	0	nsubj	\_	\_} \\ 
        \texttt{2	if	\_	\_	\_	\_	4	mark	\_	\_ \_} \\ 
        \texttt{3 Google	\_	\_	\_	\_	4	nsubj} \\ 
        \texttt{4	Morphed	\_	\_	\_	\_	0	root	\_	\_} \\ 
        \texttt{5	into	\_	\_	\_	\_	6	case	\_	\_} \\ 
        \texttt{6	GoogleOS	\_	\_	\_	\_	8	nmod	\_	\_} \\ 
        \texttt{7	?	\_	\_	\_	\_	4	punct	\_	\_}\\
        \texttt{Let me know if (...)} \\ 
        \hline \textbf{(1) Well-formatted output}\\ \hline 
        \texttt{1	What	\_	\_	\_	\_	0	nsubj	\_	\_} \\ 
        \texttt{2	if	\_	\_	\_	\_	4	mark	\_	\_ \_} \\ 
        \texttt{3 Google	\_	\_	\_	\_	4	nsubj	\_	\_} \\ 
        \texttt{4	Morphed	\_	\_	\_	\_	0	root	\_	\_} \\ 
        \texttt{5	into	\_	\_	\_	\_	6	case	\_	\_} \\ 
        \texttt{6	GoogleOS	\_	\_	\_	\_	8	nmod	\_	\_} \\ 
        \texttt{7	?	\_	\_	\_	\_	4	punct	\_	\_}\\
        \hline 
        \textbf{(2) Valid dependency tree} \\ 
        \hline 
        \texttt{1	What	\_	\_	\_	\_	4	nsubj	\_	\_} \\ 
        \texttt{2	if	\_	\_	\_	\_	4	mark	\_	\_ \_} \\ 
        \texttt{3 Google	\_	\_	\_	\_	4	nsubj	\_	\_} \\ 
        \texttt{4	Morphed	\_	\_	\_	\_	0	root	\_	\_} \\ 
        \texttt{5	into	\_	\_	\_	\_	6	case	\_	\_} \\ 
        \texttt{6	GoogleOS	\_	\_	\_	\_	4	nmod	\_	\_} \\ 
        \texttt{7	?	\_	\_	\_	\_	4	punct	\_	\_}\\
        \hline 
    \end{tabular}
    \caption{\label{fig:prompt-example-complete} Dependency parsing prompt and the resulting tree after the second post-processing step. Figure \ref{fig:prompt-example} showed the original tree.}
\end{figure}

\section{Additional results}\label{ap:experiments}

\begin{figure}[tbp]
    \centering
    \includegraphics[width=\linewidth]{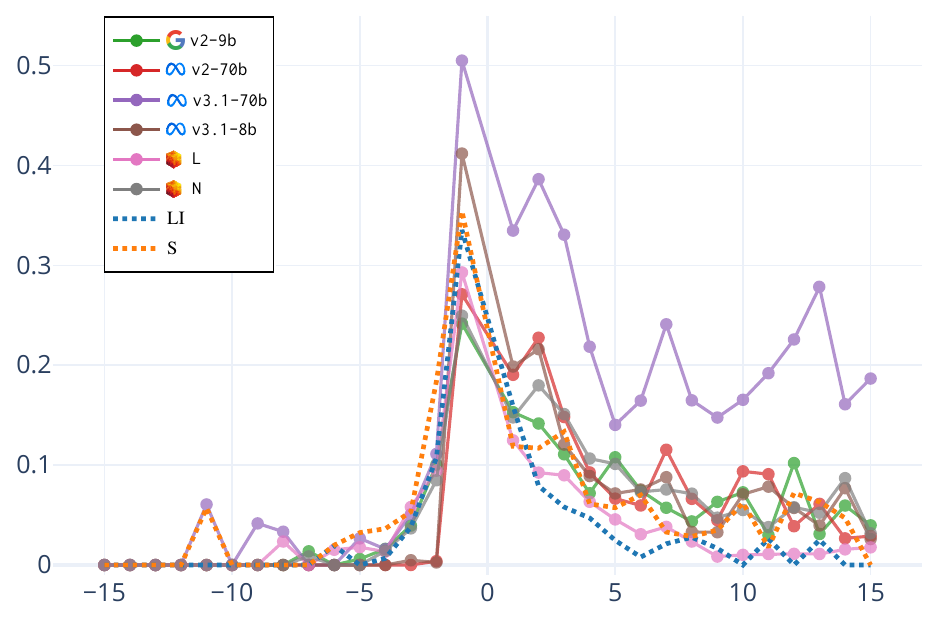}
    \caption{\label{fig:fr-displ}F-score across displacements in the French\textsubscript{GSD} test set.}
\end{figure}

\begin{figure}[tbp]
    \centering
    \includegraphics[width=\linewidth]{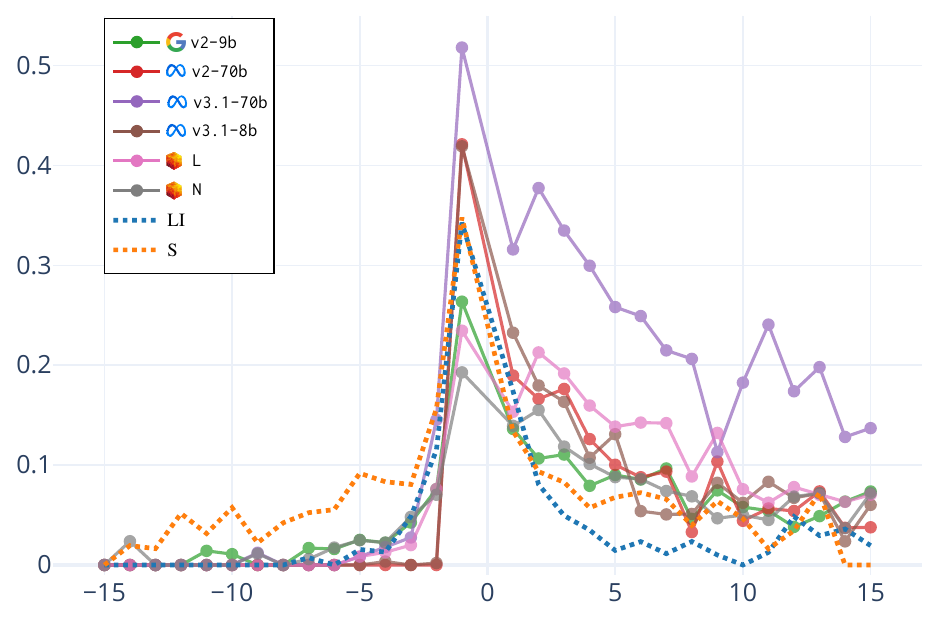}
    \caption{\label{fig:de-displ}F-score across displacements in the German\textsubscript{GSD} test set.}
\end{figure}

\begin{figure}[tbp]
    \centering
    \includegraphics[width=\linewidth]{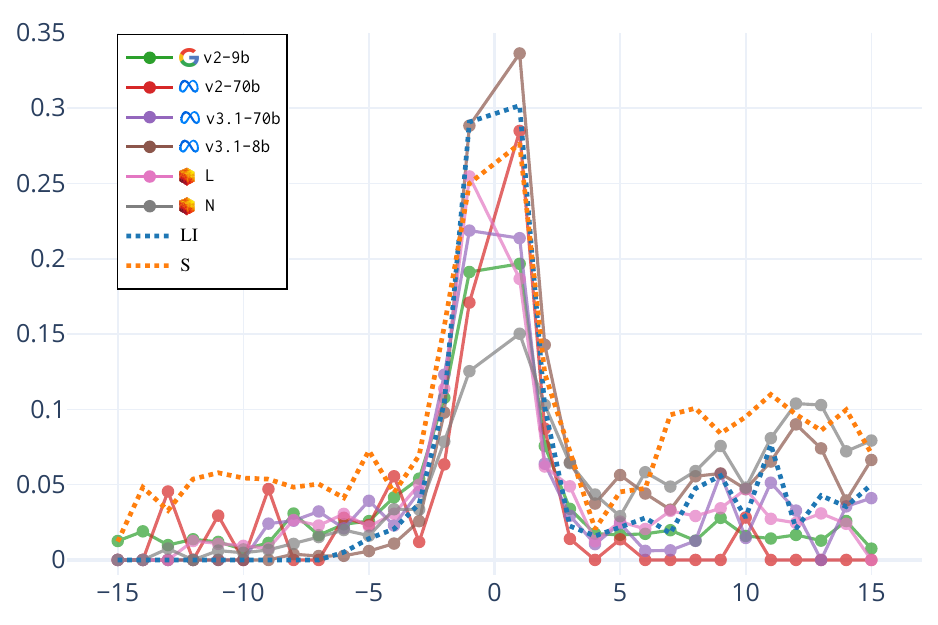}
    \caption{\label{fig:hi-displ}F-score across displacements in the Hindi\textsubscript{HDTB} test set.}
\end{figure}

Table \ref{tab:hf-models} shows the reference to each model used in our experimental study. All of them are publicly available in \href{https://huggingface.co}{HuggingFace}. Tables \ref{tab:en-pos} to \ref{tab:hi-pos} show the performance of each approach aggregating the prediction by its part-of-speech tag and Table \ref{tab:post-process-ratio} breaks down the ratio of post-processing steps performed in each experiment. Figures \ref{fig:fr-displ} to \ref{fig:hi-displ} display the 
performance on the French\textsubscript{GSD}, German\textsubscript{GSD} and Hind\textsubscript{HDBT} treebanks with respect to dependency displacement (signed dependency distance), following the definition of~\citet{anderson-gomez-rodriguez-2022-impact}, i.e., dependent index minus head index.

\begin{table}[tbp]
    \centering\scriptsize
    \setlength{\tabcolsep}{2pt}
    \begin{tabular}{p{0.3cm}p{1.2cm}|l}
        \hline 
         & \textbf{Abbrv.} & \textbf{Repository} \\
        \hline
        \multirow{4}{*}{\raisebox{-0.1\height}{\includegraphics[height=0.3cm]{images/google.pdf}}} 
        & \texttt{v1-2b}    & \href{https://huggingface.co/google/gemma-2b}{\texttt{google/gemma-2b}}\\
        & \texttt{v1-7b}    & \href{https://huggingface.co/google/gemma-7b}{\texttt{google/gemma-7b}}\\
        & \texttt{v2-9b}    & \href{https://huggingface.co/google/gemma-2-9b}{\texttt{google/gemma-2-9b}}\\\
        & \texttt{v2-27b}   & \href{https://huggingface.co/google/gemma-2-9b}{\texttt{google/gemma-2-27b}}\\ 
        \hline 
        \multirow{9}{*}{\includegraphics[height=0.22cm]{images/meta.pdf}} 
        & \texttt{v2-7b}    & \href{https://huggingface.co/meta-llama/Llama-2-7b-chat-hf}{\texttt{meta-llama/Llama-2-7b-chat-hf}}\\
        & \texttt{v2-13b}   & \href{https://huggingface.co/meta-llama/Llama-2-13b-chat-hf}{\texttt{meta-llama/Llama-2-13b-chat-hf}} \\ 
         & \texttt{v2-70b}   & \href{https://huggingface.co/meta-llama/Llama-2-70b-chat-hf}{\texttt{meta-llama/Llama-2-70b-chat-hf}} \\
        & \texttt{v3-8b}    & \href{https://huggingface.co/meta-llama/Meta-Llama-3-8B-Instruct}{\texttt{meta-llama/Meta-Llama-3-8B-Instruct}} \\
        & \texttt{v3-70b}   & \href{https://huggingface.co/meta-llama/Meta-Llama-3-70B-Instruct}{\texttt{meta-llama/Meta-Llama-3-70B-Instruct}} \\ 
        & \texttt{v3.1-8b}  & \href{https://huggingface.co/meta-llama/Llama-3.1-8B-Instruct}{\texttt{meta-llama/Llama-3.1-8B-Instruct}} \\
        & \texttt{v3.1-70b}  & \href{https://huggingface.co/meta-llama/Llama-3.1-70B-Instruct}{\texttt{meta-llama/Llama-3.1-70B-Instruct}} \\
        & \texttt{v3.2-1b}  & \href{https://huggingface.co/meta-llama/Llama-3.2-1B-Instruct}{\texttt{meta-llama/Llama-3.2-1B-Instruct}} \\
        & \texttt{v3.2-3b}  & \href{https://huggingface.co/meta-llama/Llama-3.2-3B-Instruct}{\texttt{meta-llama/Llama-3.2-3B-Instruct}} \\
        \hline 
        \multirow{7}{*}{\raisebox{-0.1\height}{\includegraphics[height=0.3cm]{images/mistral.pdf}}} 
        & \texttt{v1-7b}    & \href{https://huggingface.co/mistralai/Mistral-7B-Instruct-v0.1}{\texttt{mistralai/Mistral-7B-Instruct-v0.1}}\\
        & \texttt{v2-7b}    & \href{https://huggingface.co/mistralai/Mistral-7B-Instruct-v0.2}{\texttt{mistralai/Mistral-7B-Instruct-v0.2}}\\
        & \texttt{v3-7b}    & \href{https://huggingface.co/mistralai/Mistral-7B-Instruct-v0.3}{\texttt{mistralai/Mistral-7B-Instruct-v0.3}}\\
        & \texttt{x1-7b}    & \href{https://huggingface.co/mistralai/Mixtral-8x7B-Instruct-v0.1}{\texttt{mistralai/Mixtral-8x7B-Instruct-v0.1}} \\ 
        & \texttt{x1-22b}   & \href{https://huggingface.co/mistralai/Mixtral-8x7B-Instruct-v0.1}{\texttt{mistralai/Mixtral-8x22B-Instruct-v0.1}} \\ 
        & \texttt{nemo}     & \href{https://huggingface.co/mistralai/Mistral-Nemo-Instruct-2407}{\texttt{mistralai/Mistral-Nemo-Instruct-2407}} \\ 
        & \texttt{large}     & \href{https://huggingface.co/mistralai/Mistral-Large-Instruct-2407}{\texttt{mistralai/Mistral-Large-Instruct-2407}} \\ 
        \hline 
    \end{tabular}
    \caption{\label{tab:hf-models}\href{https://huggingface.co}{HuggingFace} reference to the instruction-based models used in our experiments.}
\end{table}

\begin{table*}[tbp]
    \setlength{\tabcolsep}{2pt}
    \centering\scriptsize
        \begin{tabular}{|p{0.3cm}p{1cm}|ccccccccccccccccc|}
        \cline{3-19} 
        \multicolumn{2}{c|}{}& ADJ & ADP & ADV & AUX & CCONJ & DET & INTJ & NOUN & NUM & PART & PRON & PROPN &  PUNCT & SCONJ & SYM & VERB & X \\
        \hline 
        & A    & 9.45 & 6.05 & 7.74 & 6.35 & 5.16 & 6.96 & 25.00 & 8.56 & 10.70 & 6.32 & 6.43 & 13.29 & 9.27 & 3.65 & 16.51 & 6.33 & 5.13 \\
        & L    & 6.26 & 5.90 & 14.80 & 1.94 & 0.14 & 0.58 & 24.17 & 6.16 & 22.14 & 11.56 & 13.47 & 23.11 & 21.12 & 1.04 & 14.68 & 8.41 & 53.85 \\
        & R    & 51.06 & 36.28 & 43.62 & 44.72 & 46.68 & 55.67 & 30.00 & 16.00 & 26.57 & 73.50 & 37.34 & 22.29 & 15.28 & 16.67 & 25.69 & 2.19 & 2.56 \\
        & RD   & 8.56 & 6.29 & 8.84 & 6.35 & 5.70 & 6.06 & 18.33 & 8.53 & 9.41 & 6.63 & 7.50 & 12.37 & 10.50 & 3.39 & 21.10 & 6.79 & 20.51 \\
        & RD*  & 8.95 & 6.00 & 8.42 & 6.74 & 5.56 & 5.80 & 21.67 & 7.88 & 11.62 & 7.70 & 6.43 & 13.00 & 10.66 & 4.95 & 11.93 & 6.18 & 10.26 \\
        & LI   & 23.94 & 15.88 & 25.00 & 22.62 & 14.79 & 22.83 & 30.83 & 12.11 & 22.14 & 29.74 & 24.85 & 20.65 & 15.12 & 11.72 & 11.93 & 7.68 & 25.64 \\
        & LI*  & 25.11 & 19.03 & 25.26 & 23.59 & 19.13 & 26.20 & 25.83 & 11.34 & 19.56 & 28.97 & 27.12 & 17.62 & 14.41 & 9.11 & 14.68 & 3.72 & 15.38 \\
        & S    &  24.94 & 23.70 & 20.49 & 24.11 & 22.25 & 24.78 & 35.83 & 14.74 & 17.71 & 27.43 & 21.52 & 21.52 & 18.38 & 11.72 & 18.35 & 9.75 & 17.95 \\
        \hline 
        \multirow{3}{*}{\raisebox{-0.1\height}{\includegraphics[height=0.3cm]{images/google.pdf}}} 
        & \texttt{v1-2b}    &  9.45 & 7.23 & 8.16 & 9.92 & 6.24 & 8.43 & 15.83 & 7.64 & 11.07 & 10.94 & 9.02 & 12.18 & 9.04 & 5.47 & 6.42 & 8.18 & 10.26 \\
        & \texttt{v1-7b}    & 10.40 & 6.59 & 9.35 & 10.95 & 7.19 & 9.17 & 14.17 & 7.83 & 8.12 & 10.48 & 11.61 & 9.82 & 9.63 & 7.81 & 12.84 & 8.45 & 5.13 \\
        & \texttt{v2-9b}    &  15.10 & 7.62 & 11.14 & 10.24 & 8.01 & 10.86 & 16.67 & 10.95 & 7.38 & 8.94 & 9.90 & 14.16 & 6.75 & 4.69 & 9.17 & 19.65 & 2.56 \\
        \hline 
        \multirow{9}{*}{\includegraphics[height=0.22cm]{images/meta.pdf}} 
        & \texttt{v2-7b}    & 8.22 & 4.57 & 10.54 & 12.70 & 3.80 & 5.69 & 10.00 & 8.34 & 4.06 & 9.71 & 8.75 & 9.39 & 6.94 & 3.91 & 8.26 & 14.55 & 2.56 \\
        & \texttt{v2-13b}   & 10.46 & 1.92 & 6.46 & 5.57 & 1.49 & 2.11 & 15.00 & 10.80 & 2.40 & 5.39 & 3.93 & 14.68 & 6.23 & 2.34 & 6.42 & 27.06 & 0.00 \\
        & \texttt{v2-70b}   & 10.07 & 7.42 & 14.88 & 14.32 & 4.34 & 7.64 & 10.00 & 12.79 & 4.80 & 14.33 & 12.36 & 10.21 & 10.14 & 5.73 & 4.59 & 13.90 & 7.69 \\
        & \texttt{v3-8b}    & 9.56 & 5.31 & 11.56 & 12.70 & 4.21 & 6.54 & 16.67 & 9.86 & 4.43 & 11.40 & 9.81 & 10.83 & 10.59 & 3.65 & 9.17 & 13.28 & 10.26 \\
        & \texttt{v3-70b}   &  36.13 & 13.82 & 38.69 & 16.59 & 8.68 & 38.22 & 42.50 & 30.41 & 11.62 & 36.52 & 33.83 & 27.15 & 24.64 & 8.33 & 15.60 & 30.67 & 17.95 \\
        & \texttt{v3.1-8b}  &  16.00 & 16.08 & 25.17 & 12.96 & 23.20 & 19.93 & 27.50 & 15.93 & 9.78 & 20.65 & 21.29 & 24.03 & 17.44 & 7.29 & 19.27 & 14.40 & 30.77 \\
        & \texttt{v3.1-70b} &  42.28 & 20.26 & 43.45 & 23.46 & 21.98 & 40.48 & 45.83 & 29.68 & 11.44 & 51.00 & 37.90 & 28.50 & 28.97 & 11.46 & 16.51 & 30.63 & 12.82 \\
        & \texttt{v3.2-1b} & 7.94 & 5.56 & 11.56 & 12.96 & 5.16 & 6.48 & 13.33 & 8.22 & 5.17 & 10.32 & 9.53 & 9.87 & 9.27 & 5.47 & 11.01 & 9.71 & 10.26 \\ 
        & \texttt{v3.2-3b} & 8.22 & 3.83 & 9.10 & 9.53 & 3.26 & 4.53 & 20.83 & 9.02 & 8.12 & 7.86 & 7.91 & 12.61 & 11.18 & 3.39 & 13.76 & 9.52 & 10.26 \\ 
        \hline 
        \multirow{8}{*}{\raisebox{-0.1\height}{\includegraphics[height=0.3cm]{images/mistral.pdf}}} 
        & \texttt{v1-7b}    &  10.85 & 4.77 & 8.50 & 9.40 & 4.48 & 5.06 & 20.00 & 9.79 & 4.61 & 8.17 & 8.65 & 11.94 & 6.17 & 2.34 & 5.50 & 16.47 & 5.13 \\
        & \texttt{v2-7b}    & 21.09 & 16.91 & 16.33 & 18.02 & 16.69 & 24.30 & 22.50 & 13.85 & 8.67 & 19.72 & 17.95 & 17.38 & 10.53 & 11.98 & 13.76 & 11.09 & 5.13 \\
        & \texttt{v3-7b}    & 30.76 & 24.68 & 27.81 & 27.28 & 24.42 & 30.89 & 15.00 & 13.85 & 8.67 & 37.60 & 21.84 & 17.00 & 13.15 & 11.72 & 19.27 & 9.79 & 12.82 \\
        & \texttt{x1-7b}    &  13.40 & 10.75 & 13.53 & 12.57 & 12.25 & 12.68 & 15.22 & 9.15 & 10.70 & 15.57 & 14.64 & 11.87 & 8.39 & 4.50 & 11.96 & 10.53 & 12.90 \\
        & \texttt{x1-22b} & 27.57 & 18.78 & 31.21 & 22.88 & 18.72 & 29.20 & 30.83 & 18.01 & 13.10 & 31.12 & 24.62 & 23.06 & 18.86 & 14.58 & 18.35 & 16.51 & 28.21\\ 
        & \texttt{nemo}     & 13.42 & 7.62 & 10.29 & 9.07 & 7.33 & 9.80 & 20.00 & 11.31 & 6.27 & 8.17 & 9.16 & 15.50 & 8.59 & 3.65 & 9.17 & 19.23 & 20.51 \\
        & \texttt{large}    &  21.48 & 5.31 & 15.99 & 9.27 & 6.78 & 10.12 & 25.83 & 18.27 & 10.52 & 10.48 & 11.01 & 22.82 & 11.95 & 2.60 & 19.27 & 38.85 & 15.38 \\
    \hline 
    \end{tabular}
    \caption{\label{tab:en-pos}UAS aggregated by universal part-of-speech tag in the English\textsubscript{EWT} test set.}
\end{table*}

\begin{table*}[tbp]
    \setlength{\tabcolsep}{2pt}
    \centering\scriptsize
        \begin{tabular}{|p{0.3cm}p{1cm}|cccccccccccccccc|}
        \cline{3-18} 
        \multicolumn{2}{c|}{}& ADJ & ADP & ADV & AUX & CCONJ & DET & INTJ & NOUN & NUM & PRON & PROPN & PUNCT & SCONJ & SYM & VERB & X \\
        \hline 
        & A    & 4.27 & 3.65 & 3.70 & 5.01 & 3.61 & 3.85 & 22.22 & 4.50 & 3.98 & 4.83 & 2.65 & 4.47 & 1.56 & 0.00 & 4.14 & 0.00 \\
        & L    & 51.23 & 1.08 & 23.82 & 0.84 & 0.00 & 0.81 & 22.22 & 5.57 & 19.03 & 4.47 & 25.92 & 11.80 & 10.94 & 5.13 & 9.62 & 42.86 \\
        & R    & 19.38 & 36.49 & 36.55 & 55.43 & 29.72 & 89.79 & 0.00 & 1.61 & 44.25 & 39.00 & 2.24 & 15.09 & 3.91 & 23.08 & 0.00 & 3.57 \\
        & RD   & 4.60 & 4.53 & 4.93 & 4.18 & 5.22 & 3.92 & 0.00 & 4.39 & 3.98 & 5.01 & 5.71 & 4.97 & 1.56 & 2.56 & 4.38 & 3.57 \\
        & RD*  &  3.28 & 4.46 & 5.54 & 2.51 & 2.41 & 3.85 & 11.11 & 4.18 & 3.10 & 4.11 & 4.29 & 4.47 & 3.91 & 0.00 & 5.36 & 3.57 \\
        & LI   &  28.41 & 15.74 & 23.41 & 21.73 & 11.65 & 30.22 & 0.00 & 5.47 & 20.35 & 20.21 & 13.67 & 10.46 & 5.47 & 15.38 & 5.24 & 17.86 \\
        & LI*  &  27.09 & 18.38 & 22.18 & 23.40 & 14.46 & 35.63 & 0.00 & 4.98 & 25.22 & 21.11 & 14.90 & 11.21 & 6.25 & 15.38 & 4.26 & 28.57 \\
        & S    & 13.79 & 24.73 & 21.15 & 26.46 & 18.88 & 34.28 & 0.00 & 9.00 & 21.24 & 21.11 & 9.39 & 12.56 & 7.81 & 15.38 & 7.43 & 14.29 \\
        \hline 
        \multirow{3}{*}{\raisebox{-0.1\height}{\includegraphics[height=0.3cm]{images/google.pdf}}} 
        & \texttt{v1-2b}    & 10.67 & 6.82 & 11.50 & 10.58 & 6.43 & 12.24 & 0.00 & 5.79 & 12.39 & 8.59 & 9.39 & 6.75 & 3.91 & 10.26 & 9.01 & 10.71 \\
        & \texttt{v1-7b}    &  10.84 & 9.05 & 11.91 & 9.47 & 5.62 & 13.52 & 0.00 & 4.93 & 9.29 & 9.84 & 7.35 & 5.99 & 3.12 & 5.13 & 7.92 & 14.29 \\
        & \texttt{v2-9b}    & 16.42 & 9.86 & 11.29 & 11.42 & 7.23 & 19.27 & 11.11 & 9.38 & 3.10 & 11.45 & 8.57 & 6.32 & 2.34 & 7.69 & 17.78 & 3.57 \\
        \hline 
        \multirow{9}{*}{\includegraphics[height=0.22cm]{images/meta.pdf}} 
        & \texttt{v2-7b}    &  10.67 & 16.15 & 17.25 & 26.46 & 13.25 & 40.70 & 22.22 & 4.72 & 19.91 & 22.00 & 5.10 & 8.77 & 2.34 & 17.95 & 2.56 & 0.00 \\
        & \texttt{v2-13b}   &9.85 & 11.82 & 17.45 & 14.21 & 8.84 & 27.86 & 0.00 & 5.31 & 11.50 & 15.56 & 5.10 & 8.35 & 3.12 & 10.26 & 5.72 & 3.57 \\
        & \texttt{v2-70b}   &  30.71 & 8.18 & 21.77 & 6.69 & 3.21 & 24.95 & 11.11 & 17.42 & 6.64 & 15.38 & 21.84 & 8.26 & 6.25 & 12.82 & 19.37 & 21.43 \\
        & \texttt{v3-8b}    &  13.79 & 12.30 & 16.43 & 17.55 & 6.83 & 28.06 & 11.11 & 4.02 & 20.35 & 17.17 & 9.18 & 9.02 & 1.56 & 10.26 & 7.67 & 3.57 \\
        & \texttt{v3-70b}   &  41.05 & 19.73 & 28.34 & 8.64 & 6.43 & 47.73 & 11.11 & 26.74 & 9.73 & 22.00 & 34.08 & 15.01 & 5.47 & 5.13 & 21.92 & 28.57 \\
        & \texttt{v3.1-8b}  & 32.02 & 17.30 & 28.95 & 25.63 & 22.49 & 53.75 & 11.11 & 11.36 & 13.72 & 25.04 & 11.02 & 13.58 & 3.12 & 12.82 & 14.86 & 7.14 \\
        & \texttt{v3.1-70b} & 53.20 & 27.91 & 37.78 & 13.93 & 14.46 & 59.77 & 0.00 & 27.49 & 15.49 & 28.09 & 26.12 & 23.44 & 5.47 & 15.38 & 29.23 & 21.43 \\
        & \texttt{v3.2-1b} & 7.22 & 5.41 & 6.16 & 3.90 & 3.61 & 9.87 & 11.11 & 5.47 & 8.85 & 8.77 & 4.69 & 4.81 & 3.91 & 12.82 & 10.96 & 0.00 \\
        & \texttt{v3.2-3b} & 10.84 & 5.81 & 8.21 & 9.75 & 2.41 & 12.24 & 11.11 & 7.18 & 6.64 & 7.51 & 10.00 & 7.00 & 3.91 & 5.13 & 9.01 & 14.29  \\
        \hline 
        \multirow{7}{*}{\raisebox{-0.1\height}{\includegraphics[height=0.3cm]{images/mistral.pdf}}} 
        & \texttt{v1-7b}    & 11.82 & 6.69 & 10.06 & 4.46 & 4.42 & 12.17 & 0.00 & 6.22 & 6.19 & 8.94 & 9.80 & 5.90 & 2.34 & 7.69 & 14.98 & 14.29 \\
        & \texttt{v2-7b}    &21.35 & 19.05 & 17.45 & 14.48 & 12.05 & 33.60 & 0.00 & 11.74 & 7.96 & 19.32 & 11.43 & 10.12 & 3.12 & 20.51 & 12.55 & 21.43 \\
        & \texttt{v3-7b}    & 26.44 & 29.39 & 31.62 & 31.20 & 21.29 & 70.05 & 11.11 & 10.40 & 16.37 & 29.34 & 13.27 & 12.14 & 3.12 & 17.95 & 11.21 & 14.29 \\
        & \texttt{x1-7b}    & 16.09 & 6.42 & 9.03 & 6.69 & 4.02 & 15.89 & 22.22 & 10.93 & 6.64 & 7.69 & 7.76 & 6.16 & 5.47 & 5.13 & 27.65 & 0.00 \\
        & \texttt{x1-22b} & 29.23 & 20.61 & 26.28 & 14.76 & 17.27 & 38.54 & 22.22 & 14.47 & 18.58 & 21.29 & 17.76 & 13.24 & 7.03 & 23.08 & 14.98 & 7.14 \\
        & \texttt{nemo}     & 17.24 & 10.34 & 14.17 & 8.64 & 6.43 & 20.76 & 11.11 & 11.95 & 4.42 & 11.63 & 13.88 & 8.94 & 3.91 & 5.13 & 19.12 & 10.71 \\
        & \texttt{large}    &  12.97 & 13.38 & 14.99 & 16.43 & 11.24 & 25.42 & 0.00 & 5.63 & 16.37 & 16.28 & 8.37 & 8.60 & 5.47 & 7.69 & 6.21 & 7.14 \\
    \hline 
    \end{tabular}
    \caption{\label{tab:fr-pos}UAS aggregated by universal part-of-speech tag in the French\textsubscript{GSD} test set.}
\end{table*}

\begin{table*}[thpb!]
    \setlength{\tabcolsep}{2pt}
    \centering\scriptsize
        \begin{tabular}{|p{0.3cm}p{1cm}|ccccccccccccccccc|}
        \cline{3-19} 
        \multicolumn{2}{c|}{} & ADJ & ADP & ADV & AUX & CCONJ & DET & INTJ & NOUN & NUM & PART & PRON & PROPN & PUNCT & SCONJ & SYM & VERB & X \\
        \hline 
        & A    & 5.36 & 4.61 & 4.36 & 5.51 & 3.46 & 5.26 & 0.00 & 4.85 & 6.01 & 5.24 & 3.83 & 6.26 & 4.52 & 4.35 & 0.00 & 4.07 & 16.00 \\
        & L    & 1.75 & 1.18 & 12.31 & 23.77 & 3.68 & 0.80 & 25.00 & 2.28 & 9.87 & 10.48 & 19.72 & 30.04 & 23.13 & 0.00 & 25.00 & 1.58 & 28.00 \\
        & R    & 69.43 & 29.95 & 34.35 & 6.23 & 43.29 & 68.97 & 0.00 & 15.49 & 63.52 & 47.62 & 17.16 & 8.51 & 10.49 & 1.86 & 50.00 & 2.64 & 24.00 \\
        & RD   & 4.77 & 5.73 & 6.85 & 5.80 & 6.49 & 4.95 & 25.00 & 6.33 & 4.72 & 4.76 & 5.25 & 6.36 & 6.22 & 5.59 & 0.00 & 6.18 & 16.00 \\
        & RD*  &   6.33 & 5.04 & 5.92 & 6.23 & 5.84 & 5.31 & 0.00 & 6.62 & 4.29 & 5.71 & 5.39 & 7.44 & 5.88 & 8.07 & 0.00 & 5.35 & 12.00 \\
        & LI   & 26.39 & 15.13 & 18.07 & 16.23 & 16.88 & 28.69 & 25.00 & 9.58 & 34.33 & 11.43 & 15.60 & 18.10 & 17.25 & 1.24 & 0.00 & 5.05 & 24.00 \\
        & LI*  & 27.85 & 15.75 & 22.12 & 17.97 & 16.67 & 31.83 & 0.00 & 8.39 & 31.76 & 20.00 & 20.71 & 18.59 & 14.08 & 1.86 & 0.00 & 3.77 & 20.00 \\
        & S    &  29.11 & 18.56 & 18.69 & 12.03 & 16.45 & 26.57 & 0.00 & 12.18 & 24.46 & 25.71 & 15.60 & 10.67 & 11.63 & 3.73 & 25.00 & 7.69 & 20.00 \\
        \hline 
        \multirow{3}{*}{\raisebox{-0.1\height}{\includegraphics[height=0.3cm]{images/google.pdf}}} 
        & \texttt{v1-2b}    &17.33 & 10.46 & 13.71 & 6.23 & 9.96 & 20.78 & 0.00 & 7.68 & 17.60 & 10.00 & 11.91 & 9.39 & 9.43 & 4.97 & 0.00 & 6.93 & 8.00 \\
        & \texttt{v1-7b}    & 26.29 & 11.71 & 14.95 & 3.33 & 16.02 & 27.94 & 0.00 & 8.07 & 11.16 & 16.67 & 9.50 & 7.63 & 6.22 & 0.00 & 50.00 & 20.12 & 12.00 \\
        & \texttt{v2-9b}    & 18.31 & 10.40 & 12.85 & 5.65 & 8.66 & 18.52 & 0.00 & 7.39 & 8.15 & 13.33 & 8.09 & 10.67 & 5.96 & 3.11 & 25.00 & 22.76 & 16.00 \\
        \hline 
        \multirow{9}{*}{\includegraphics[height=0.22cm]{images/meta.pdf}} 
        & \texttt{v2-7b}    &  32.23 & 16.56 & 19.39 & 6.38 & 21.21 & 35.15 & 0.00 & 10.03 & 25.75 & 17.62 & 12.77 & 9.00 & 9.22 & 1.24 & 50.00 & 4.90 & 4.00 \\
        & \texttt{v2-13b}   & 34.54 & 18.22 & 23.13 & 5.58 & 18.94 & 36.85 & 0.00 & 10.17 & 28.10 & 18.10 & 12.19 & 9.81 & 9.32 & 0.00 & 0.00 & 4.35 & 15.38 \\
        & \texttt{v2-70b}   & 42.16 & 18.24 & 28.12 & 11.74 & 21.43 & 46.82 & 0.00 & 14.01 & 12.88 & 24.29 & 18.44 & 18.49 & 13.83 & 1.24 & 0.00 & 13.26 & 28.00 \\
        & \texttt{v3-8b}    & 58.13 & 25.34 & 29.36 & 6.96 & 30.74 & 58.49 & 0.00 & 13.47 & 23.18 & 29.52 & 18.72 & 11.94 & 13.95 & 1.24 & 50.00 & 17.41 & 28.00 \\
        & \texttt{v3-70b}   &  48.69 & 24.22 & 31.78 & 7.25 & 11.26 & 47.92 & 0.00 & 25.81 & 18.45 & 32.38 & 24.26 & 29.75 & 17.93 & 1.86 & 0.00 & 30.22 & 32.00 \\
        & \texttt{v3.1-8b}  &45.86 & 20.55 & 28.35 & 9.28 & 27.92 & 47.44 & 0.00 & 13.79 & 17.17 & 30.00 & 21.70 & 20.35 & 15.43 & 1.86 & 25.00 & 13.79 & 28.00 \\
        & \texttt{v3.1-70b} &  60.76 & 26.40 & 34.19 & 16.38 & 23.81 & 53.80 & 25.00 & 27.19 & 20.60 & 39.52 & 23.40 & 27.50 & 24.36 & 1.86 & 0.00 & 27.51 & 24.00 \\
        & \texttt{v3.2-1b} & 10.42 & 6.54 & 9.66 & 6.38 & 5.41 & 14.32 & 0.00 & 6.88 & 10.30 & 7.14 & 12.34 & 8.02 & 8.54 & 2.48 & 25.00 & 10.85 & 12.00 \\
        & \texttt{v3.2-3b} & 30.19 & 15.88 & 18.07 & 6.81 & 21.86 & 28.65 & 25.00 & 10.77 & 23.61 & 20.00 & 12.62 & 10.37 & 8.54 & 2.48 & 25.00 & 9.27 & 20.00 \\
        \hline 
        \multirow{8}{*}{\raisebox{-0.1\height}{\includegraphics[height=0.3cm]{images/mistral.pdf}}} 
        & \texttt{v1-7b}    &  21.71 & 13.14 & 13.24 & 4.35 & 14.29 & 23.43 & 0.00 & 8.13 & 9.44 & 11.43 & 11.77 & 10.27 & 7.23 & 1.24 & 0.00 & 16.28 & 20.00 \\
        & \texttt{v2-7b}    &33.79 & 18.12 & 21.81 & 4.64 & 19.48 & 35.06 & 0.00 & 11.06 & 15.45 & 18.57 & 12.77 & 13.21 & 9.30 & 1.24 & 0.00 & 14.32 & 12.00 \\
        & \texttt{v3-7b}    &62.51 & 27.90 & 31.31 & 7.97 & 37.88 & 61.54 & 0.00 & 13.95 & 24.89 & 37.62 & 18.01 & 10.47 & 11.59 & 0.62 & 25.00 & 8.89 & 12.00 \\
        & \texttt{x1-7b}    & 11.11 & 0.00 & 30.00 & 8.33 & 0.00 & 30.43 &  & 13.51 & 0.00 & 0.00 & 22.22 & 0.00 & 18.75 & 0.00 &  & 0.00 & 0.00 \\
        & \texttt{x2-22b} & 30.19 & 18.80 & 21.88 & 9.42 & 16.23 & 32.40 & 0.00 & 13.02 & 19.31 & 20.48 & 19.29 & 17.03 & 13.49 & 3.11 & 0.00 & 13.34 & 32.00 \\
        & \texttt{nemo}     &  11.88 & 6.54 & 10.20 & 5.36 & 6.93 & 13.04 & 0.00 & 9.03 & 4.29 & 9.52 & 6.81 & 11.06 & 6.43 & 0.62 & 25.00 & 27.51 & 20.00 \\
        & \texttt{large}    &  21.81 & 7.60 & 11.99 & 4.78 & 6.49 & 14.90 & 25.00 & 12.63 & 6.87 & 11.90 & 8.09 & 12.23 & 10.23 & 0.62 & 25.00 & 36.17 & 24.00 \\
    \hline 
    \end{tabular}
    \caption{\label{tab:de-pos}UAS aggregated by universal part-of-speech tag in the German\textsubscript{GSD} test set.}
\end{table*}

\begin{table*}[tbp]
    \setlength{\tabcolsep}{2pt}
    \centering\scriptsize
        \begin{tabular}{|p{0.3cm}p{1cm}|ccccccccccccccc|}
        \cline{3-17} 
        \multicolumn{2}{c|}{} & ADJ & ADP & ADV & AUX & CCONJ & DET & NOUN & NUM & PART & PRON & PROPN & PUNCT & SCONJ & VERB & X \\
        \hline 
        & A    & 4.56 & 4.88 & 4.93 & 4.77 & 5.83 & 5.77 & 4.99 & 2.89 & 6.06 & 5.39 & 4.44 & 5.41 & 2.29 & 4.48 & 11.11 \\
        & L    & 6.34 & 83.96 & 0.66 & 68.81 & 0.00 & 0.00 & 0.09 & 0.00 & 40.32 & 0.29 & 0.09 & 16.78 & 0.76 & 4.97 & 11.11 \\
        & R    & 73.13 & 0.07 & 22.37 & 0.13 & 27.09 & 83.09 & 25.65 & 75.32 & 28.06 & 32.87 & 31.70 & 13.55 & 0.00 & 5.03 & 55.56 \\
        & RD   & 5.40 & 4.58 & 3.29 & 5.73 & 3.78 & 4.70 & 4.99 & 4.76 & 5.47 & 5.90 & 4.15 & 4.96 & 3.82 & 4.97 & 0.00 \\
        & RD*  &  4.84 & 4.68 & 3.29 & 5.64 & 5.51 & 6.44 & 4.68 & 4.91 & 5.02 & 5.76 & 4.19 & 5.04 & 3.97 & 5.66 & 0.00 \\
        & LI   &28.98 & 27.87 & 8.55 & 38.46 & 10.08 & 40.13 & 11.14 & 30.30 & 25.11 & 15.89 & 16.34 & 17.02 & 0.76 & 7.08 & 22.22 \\
        & LI*  &30.20 & 28.24 & 9.87 & 32.94 & 12.44 & 37.85 & 12.12 & 27.99 & 23.93 & 16.69 & 15.93 & 13.22 & 0.92 & 7.71 & 33.33 \\
        & S    & 23.91 & 23.91 & 11.84 & 27.97 & 10.71 & 24.83 & 13.80 & 19.19 & 19.79 & 15.52 & 13.36 & 16.20 & 4.89 & 11.70 & 11.11 \\
        \hline 
        \multirow{3}{*}{\raisebox{-0.1\height}{\includegraphics[height=0.3cm]{images/google.pdf}}} 
        & \texttt{v1-2b}    & 9.77 & 23.07 & 5.26 & 15.43 & 3.15 & 12.62 & 4.74 & 8.23 & 13.59 & 6.78 & 6.35 & 8.02 & 1.83 & 5.58 & 11.11 \\
        & \texttt{v1-7b}    & 16.16 & 8.24 & 6.25 & 4.85 & 6.77 & 21.34 & 6.26 & 11.69 & 8.12 & 8.82 & 8.54 & 4.71 & 0.00 & 15.75 & 0.00 \\
        & \texttt{v2-9b}    & 13.76 & 17.97 & 7.89 & 13.34 & 5.35 & 16.11 & 6.63 & 11.54 & 12.11 & 8.82 & 8.18 & 4.30 & 1.22 & 18.00 & 22.22 \\
        \hline 
        \multirow{9}{*}{\includegraphics[height=0.22cm]{images/meta.pdf}} 
        & \texttt{v2-7b}    &  14.51 & 12.82 & 5.92 & 11.08 & 5.51 & 21.34 & 5.86 & 12.41 & 10.49 & 9.55 & 9.08 & 7.81 & 1.37 & 5.14 & 0.00 \\
        & \texttt{v2-13b}   & 9.91 & 32.48 & 4.61 & 22.66 & 2.68 & 18.12 & 3.96 & 7.22 & 16.10 & 9.91 & 6.17 & 9.34 & 1.83 & 5.26 & 0.00 \\
        & \texttt{v2-70b}   &  10.77 & 28.76 & 7.32 & 20.89 & 4.60 & 22.68 & 3.77 & 10.00 & 19.72 & 9.14 & 8.22 & 10.45 & 0.00 & 3.94 &  \\
        & \texttt{v3-8b}    & 20.81 & 28.20 & 13.16 & 23.91 & 5.51 & 28.99 & 6.97 & 12.70 & 24.52 & 11.88 & 8.29 & 6.86 & 0.15 & 16.32 & 11.11 \\
        & \texttt{v3-70b}   &  20.61 & 16.79 & 11.51 & 13.01 & 8.92 & 22.76 & 10.29 & 17.99 & 16.13 & 14.98 & 12.95 & 8.22 & 1.03 & 5.70 & 0.00 \\
        & \texttt{v3.1-8b}  & 25.27 & 30.57 & 12.50 & 32.86 & 7.87 & 38.79 & 9.83 & 22.37 & 20.09 & 17.42 & 11.69 & 7.02 & 0.00 & 13.55 & 11.11 \\
        & \texttt{v3.1-70b} & 17.67 & 18.19 & 6.47 & 15.46 & 11.52 & 22.44 & 10.38 & 12.95 & 18.28 & 13.69 & 12.79 & 9.01 & 2.40 & 4.19 & 0.00 \\
        & \texttt{v3.2-1b} & 8.08 & 6.87 & 2.96 & 6.19 & 5.98 & 11.14 & 4.23 & 6.49 & 5.17 & 8.75 & 5.36 & 6.40 & 1.07 & 7.25 & 0.00 \\
        & \texttt{v3.2-3b} & 7.56 & 35.68 & 5.92 & 22.49 & 1.26 & 8.46 & 2.86 & 3.46 & 16.25 & 4.74 & 4.53 & 7.07 & 1.07 & 8.84 & 11.11 \\
        \hline 
        \multirow{8}{*}{\raisebox{-0.1\height}{\includegraphics[height=0.3cm]{images/mistral.pdf}}} 
        & \texttt{v1-7b}    &  9.53 & 19.55 & 4.28 & 6.44 & 2.68 & 13.42 & 3.68 & 5.63 & 9.16 & 7.14 & 6.08 & 2.15 & 0.15 & 17.74 & 0.00 \\
        & \texttt{v2-7b}    & 15.88 & 22.49 & 11.51 & 9.49 & 4.88 & 22.01 & 7.14 & 10.53 & 17.58 & 8.45 & 8.29 & 3.88 & 0.46 & 18.64 & 0.00 \\
        & \texttt{v3-7b}    &  29.64 & 30.29 & 16.12 & 18.31 & 8.98 & 44.56 & 11.01 & 26.84 & 21.42 & 14.58 & 14.98 & 6.16 & 0.46 & 16.73 & 22.22 \\
        & \texttt{x1-7b}    & 29.64 & 30.29 & 16.12 & 18.31 & 8.98 & 44.56 & 11.01 & 26.98 & 21.42 & 14.58 & 14.96 & 6.16 & 0.46 & 16.73 & 22.22 \\
        & \texttt{x1-22b} & 28.64 & 33.08 & 9.83 & 20.04 & 13.70 & 33.33 & 11.79 & 27.62 & 17.60 & 16.85 & 15.19 & 10.49 & 1.53 & 9.16 & 16.67 \\
        & \texttt{nemo}     &  9.58 & 11.84 & 4.61 & 7.90 & 3.31 & 10.07 & 4.99 & 7.36 & 11.37 & 4.96 & 5.16 & 5.12 & 0.92 & 20.69 & 11.11 \\
        & \texttt{large}    &  25.37 & 14.66 & 11.51 & 9.14 & 12.64 & 29.81 & 10.38 & 21.94 & 16.49 & 13.20 & 13.77 & 8.02 & 1.03 & 5.43 & 0.00 \\
    \hline 
    \end{tabular}
    \caption{\label{tab:hi-pos}UAS aggregated by universal part-of-speech tag in the Hindi\textsubscript{HDTB} test set.}
\end{table*}

\begin{table*}[tbp]
    \setlength{\tabcolsep}{2pt}
    \centering\footnotesize
        \begin{tabular}{|p{0.3cm}p{1.3cm}|ccc|ccc|ccc|ccc|}
        \cline{3-14} 
        \multicolumn{2}{c|}{}& \multicolumn{3}{c|}{en\textsubscript{EWT}} & \multicolumn{3}{c|}{fr\textsubscript{GSD}} & \multicolumn{3}{c|}{de\textsubscript{GSD}} & \multicolumn{3}{c|}{hi\textsubscript{HDTB}}\\
        \cline{3-14}
        \multicolumn{2}{c|}{}& \textbf{NP} & \textbf{P1} & \textbf{P2} & \textbf{NP} & \textbf{P1} & \textbf{P2}  & \textbf{NP} & \textbf{P1} & \textbf{P2}  & \textbf{NP} & \textbf{P1} & \textbf{P2}\\ 
        \hline 
        \multirow{3}{*}{\raisebox{-0.1\height}{\includegraphics[height=0.3cm]{images/google.pdf}}} 
        & \texttt{v1-2b}    &  13.38 & 16.18 & 70.44 & 7.45 & 14.90 & 77.64 & 7.06 & 12.90 & 80.04 & 6.06 & 16.27 & 77.67 \\
        & \texttt{v1-7b}    &   14.44 & 16.51 & 69.04 & 6.97 & 10.82 & 82.21 & 10.75 & 0.00 & 89.25 & 8.43 & 0.12 & 91.45 \\
        & \texttt{v2-9b}    &   15.17 & 0.05 & 84.79 & 4.81 & 0.00 & 95.19 & 5.32 & 0.00 & 94.68 & 3.27 & 0.00 & 96.73 \\
        \hline 
        \multirow{9}{*}{\includegraphics[height=0.22cm]{images/meta.pdf}} 
        & \texttt{v2-7b}    &  24.22 & 0.00 & 75.78 & 1.92 & 2.64 & 95.43 & 4.09 & 8.70 & 87.21 & 5.82 & 5.23 & 88.95 \\
        & \texttt{v2-13b}   &  14.59 & 0.00 & 85.41 & 22.12 & 10.58 & 67.31 & 32.89 & 5.12 & 61.98 & 18.71 & 13.18 & 68.11 \\
        & \texttt{v2-70b}   & 46.89 & 0.00 & 53.11 & 13.70 & 0.00 & 86.30 & 23.34 & 0.10 & 76.56 & 7.47 & 8.71 & 83.82 \\
        & \texttt{v3-8b}    &   49.01 & 0.00 & 50.99 & 4.81 & 6.25 & 88.94 & 22.42 & 0.00 & 77.58 & 0.48 & 0.00 & 99.52 \\
        & \texttt{v3-70b}   & 58.69 & 0.10 & 41.21 & 33.41 & 0.00 & 66.59 & 28.25 & 0.00 & 71.75 & 24.76 & 12.72 & 62.52 \\
        & \texttt{v3.1-8b}  &  34.38 & 0.00 & 65.62 & 5.29 & 0.48 & 94.23 & 12.90 & 0.00 & 87.10 & 2.43 & 0.00 & 97.57 \\
        & \texttt{v3.1-70b} &  65.86 & 0.05 & 34.09 & 42.79 & 0.24 & 56.97 & 46.57 & 0.00 & 53.43 & 26.14 & 11.34 & 62.52 \\
        & \texttt{v3.2-1b} & 5.63 & 14.06 & 80.31 & 2.16 & 8.41 & 89.42 & 1.84 & 9.72 & 88.43 & 1.37 & 8.19 & 90.44\\
        & \texttt{v3.2-3b} & 5.15 & 14.68 & 80.16 & 2.40 & 10.58 & 87.02 & 3.99 & 9.01 & 87.00 & 2.73 & 12.17 & 85.10 \\
        \hline 
        \multirow{8}{*}{\raisebox{-0.1\height}{\includegraphics[height=0.3cm]{images/mistral.pdf}}} 
        & \texttt{v1-7b}    &  16.85 & 0.00 & 83.15 & 5.29 & 0.00 & 94.71 & 5.94 & 0.00 & 94.06 & 0.12 & 0.00 & 99.88 \\
        & \texttt{v2-7b}    &15.12 & 0.00 & 84.88 & 3.85 & 0.00 & 96.15 & 5.02 & 0.00 & 94.98 & 0.36 & 0.00 & 99.64 \\
        & \texttt{v3-7b}    & 28.17 & 0.05 & 71.79 & 11.06 & 0.00 & 88.94 & 19.55 & 0.00 & 80.45 & 0.71 & 0.00 & 99.29 \\
        & \texttt{x1-7b}    &  26.22 & 13.27 & 60.51 & 3.37 & 0.00 & 96.63 & 25.00 & 12.50 & 62.50 & 1.25 & 0.06 & 98.69 \\
        & \texttt{x1-22b}  & 41.98 & 29.22 & 28.79 & 32.69 & 26.92 & 40.38 & 37.46 & 26.41 & 36.13 & 30.88 & 33.02 & 36.10 \\ 
        & \texttt{nemo}     & 15.74 & 0.00 & 84.26 & 3.61 & 0.00 & 96.39 & 4.09 & 0.00 & 95.91 & 0.48 & 0.00 & 99.52 \\
        & \texttt{large}    & 18.25 & 0.00 & 81.75 & 22.84 & 15.14 & 62.02 & 7.88 & 0.00 & 92.12 & 18.26 & 11.20 & 70.54 \\
    \hline 
    \end{tabular}
    \caption{\label{tab:post-process-ratio}Distribution of the amount of post-processing steps performed in each zero-shot parser. \textbf{NP} represents the ratio of generated trees that did not require post-processing (only removing non-tabular lines), \textbf{P1} for those trees that only required the first post-processing step (e.g. removing extra columns) and \textbf{P2} for those trees that required of the full post-processing step (e.g. breaking cycles).}
\end{table*}

\section{Official language support}\label{ap:language}

Table \ref{tab:support_grouped_corrected} shows which of our four target languages are supported by each of the models we used, according to the official documentation provided with each model.

\begin{table*}[tbp]
    \centering\footnotesize
    \begin{tabular}{|cl|c|c|c|c|}
        \hline
        &\textbf{Model} & \textbf{English} & \textbf{French} & \textbf{German} & \textbf{Hindi} \\ 
        \hline 
        \multirow{3}{*}{\raisebox{-0.1\height}{\includegraphics[height=0.3cm]{images/google.pdf}}} 
        & \texttt{v1-2b}    & \ding{51} &  &  &  \\
        & \texttt{v1-7b}    & \ding{51} &  &  &  \\
        & \texttt{v2-9b}    & \ding{51} &  &  &  \\
        \hline
        \multirow{9}{*}{\includegraphics[height=0.22cm]{images/meta.pdf}} 
        & \texttt{v2-7b}    & \ding{51} &  &  &  \\
        & \texttt{v2-13b}   & \ding{51} &  &  &  \\
        & \texttt{v2-70b}   & \ding{51} &  &  &  \\
        & \texttt{v3-8b}    & \ding{51} & \ding{51} &  & \ding{51} \\
        & \texttt{v3-70b}   & \ding{51} & \ding{51} &  & \ding{51} \\
        & \texttt{v3.1-8b}  & \ding{51} &  \ding{51}  & \ding{51} &  \ding{51}  \\
        & \texttt{v3.1-70b} & \ding{51} &\ding{51}  & \ding{51} &  \ding{51} \\
        & \texttt{v3.3-1b} & \ding{51} &\ding{51}  & \ding{51} &  \ding{51} \\
        & \texttt{v3.3-3b} & \ding{51} &\ding{51}  & \ding{51} &  \ding{51} \\
        \hline 
        \multirow{6}{*}{\raisebox{-0.1\height}{\includegraphics[height=0.3cm]{images/mistral.pdf}}} 
        & \texttt{v1-7b}    & \ding{51} & \ding{51} & &  \\
        & \texttt{v2-7b}    & \ding{51} &  &  &  \\
        & \texttt{v3-7b}    & \ding{51} &  &  &  \\
        & \texttt{x1-7b}    & \ding{51} & \ding{51} & \ding{51} &  \\
        & \texttt{nemo}     & \ding{51} & \ding{51} & \ding{51} & \ding{51} \\
        & \texttt{large}    & \ding{51} & \ding{51} & \ding{51} & \ding{51} \\
        \hline 
    \end{tabular}
    \caption{\label{tab:support_grouped_corrected}Language support across different models. A tick symbol (\ding{51}) indicates that the model supports the respective language, while empty cells indicate lack of support.}
\end{table*}

\end{document}